\pdfoutput=1
\documentclass[11pt]{article}

\usepackage[]{acl}

\usepackage{times}
\usepackage{latexsym}
\usepackage{url}
\usepackage{graphicx, subfigure, float}
\usepackage{mathrsfs}
\usepackage{amsfonts}
\usepackage{psfrag, amssymb, amsmath}
\usepackage{color,xcolor}
\usepackage{amsmath}
\usepackage{algorithm}
\usepackage{algorithmic}
\usepackage{wrapfig}
\usepackage{comment}
\usepackage{booktabs}
\usepackage[normalem]{ulem}
\usepackage{booktabs}
\usepackage{graphicx}
\usepackage{multirow}
\usepackage{enumitem}
\usepackage{fdsymbol}

\usepackage[T1]{fontenc}
\usepackage[utf8]{inputenc}

\usepackage{microtype}

\usepackage{tablefootnote}

\title{Out-of-Distribution Generalization in Text Classification: \\Past, Present, and Future}

\author{ 
 Linyi Yang$^{\spadesuit \heartsuit}\footnotemark[1]$\hspace{0.5mm},
 Yaoxiao Song$^{\spadesuit \heartsuit}\footnotemark[1]$\hspace{0.5mm},
 Xuan Ren$^{\clubsuit}\footnotemark[1]$, Chenyang Lyu$^{\diamondsuit}$\hspace{0.5mm}, Yidong Wang$^{\spadesuit}$ \\
  \bf{Lingqiao Liu$^{\clubsuit}$\hspace{0.5mm}, Jindong Wang$^{\varheartsuit}$\hspace{0.5mm}, Jennifer Foster$^{\diamondsuit}$\hspace{0.5mm},  Yue Zhang$^{\spadesuit \heartsuit}$}\hspace{0.2mm}\hspace{1.5mm} \\
$^\spadesuit$ Westlake University \ \ \quad$^\heartsuit$ Westlake Institute for Advanced Study \ \ $^\clubsuit$University of Adelaide\\ $^\diamondsuit$Dublin City University \ \ $^\varheartsuit$ Microsoft Research Asia
 \\
 \quad\texttt{\{yanglinyi,zhangyue\}@westlake.edu.cn}\\
}

\begin{document}
\maketitle
\def\thefootnote{1}\footnotetext{These authors contributed equally to this work.}
\begin{abstract}

Machine learning (ML) systems in natural language processing (NLP) face significant challenges in generalizing to out-of-distribution (OOD) data, where the test distribution differs from the training data distribution. This poses important questions about the robustness of NLP models and their high accuracy, which may be artificially inflated due to their underlying sensitivity to systematic biases. Despite these challenges, there is a lack of comprehensive surveys on the generalization challenge from an OOD perspective in text classification. Therefore, this paper aims to fill this gap by presenting the first comprehensive review of recent progress, methods, and evaluations on this topic. We furth discuss the challenges involved and potential future research directions. By providing quick access to existing work, we hope this survey will encourage future research in this area. 

%Finally, from the OOD perspective, we analyze the existing issues and compare the development difference between NLP and CV. We also discuss the promising future directions for NLP generalization.

%Finally, from the OOD perspective, we analyze the development difference of NLP with CV, and discuss the promising future direction for NLP generalization.

\end{abstract}

\section{Introduction}\label{intro}
Pre-trained Language Models (PLMs) \cite{devlin2018bert,liu2019roberta,radford2018improving} have revolutionized natural language processing (NLP) and enabled remarkable advances in Large-scale Language Models (LLMs) \cite{touvron2023llama,gozalo2023chatgpt,pichai2023important}. Despite substantial progress in developing accurate models in several text classification tasks, including sentiment analysis \cite{kaushik2019learning,ni2019justifying,yang2021exploring,lu2022rationale}, natural language inference \cite{williams2018broad}, and machine reading comprehension \cite{kaushik2018much,sugawara2020assessing}, a major challenge persists -- out-of-distribution (OOD) generalization -- which entails the ability of a model to accurately classify text instances from distributions different from those of the training data \cite{ben2010theory,hendrycks2017baseline,hupkes2022state}. This paper aims to provide a comprehensive overview of the current state of research in OOD generalization for text classification, highlighting key methodologies, advancements, and unique challenges.

The importance of OOD generalization in text classification cannot be overstated, as real-world data often exhibit diversity and unpredictability. Numerous applications, such as sentiment analysis, document categorization, and spam detection \cite{ood,yang2022glue}, necessitate models capable of adapting to novel and unforeseen data distributions. While machine learning models generally demonstrate strong in-distribution performance, their performance frequently deteriorates when confronted with OOD instances, underscoring the need for effective strategies that facilitate generalization beyond the training distribution. Although research on OOD generalization in NLP is emerging, it is not on the scale of other tasks like computer vision \cite{ye2021ood,koh2021wilds} and time series \cite{du2021adarnn,gagnon2022woods}. Furthermore, most related surveys in NLP focus on measuring and improving model robustness against adversarial attacks \cite{schlegel2020beyond,arora2021types}, or providing causal explanations \cite{keith2020text}. Among them, \citet{wang2021measure} is the most relevant review to this paper, but their work does not differentiate between data-level variance and distorted features.

To address these limitations, this survey provides an extensive examination of the existing literature on OOD generalization in text classification, covering a diverse array of techniques and approaches. We focus on two salient components that affect OOD generalization: Systematic Data Variance and Distorted Features. Additionally, we discuss the evaluation metrics and benchmarks employed to assess the effectiveness of these techniques, as well as the limitations and drawbacks of current methodologies. 

Throughout this survey, we trace the evolution of OOD generalization techniques in text classification, from the early approaches based on traditional machine learning algorithms to more recent advancements driven by deep learning architectures, also including the discussion of the most recent emergent abilities of LLMs. We identify the key innovations and breakthroughs that have shaped the field, while also highlighting areas where progress has been relatively slow or incremental. Our analysis emphasizes the interconnected nature of these advancements and the importance of driving fundamental research in the generalization problem towards unforeseen data distributions.

In addition to providing a thorough review of existing research, this survey aims to identify open challenges and future directions for OOD generalization in text classification, especially for LLMs. We discuss the limitations of current techniques, potential avenues for improving model robustness and adaptability, and emerging research trends that may contribute to the development of more effective OOD generalization strategies. 

The remainder of this survey is organized as follows: we propose a new taxonomy towards OOD robustness, summarize relevant related work, and present an overview of current methods for addressing the OOD generalization problem in Section 2. In particular, we identify two salient components affecting the OOD generalization, namely \emph{Systematic Data Variance} and \emph{Distorted Features}. We review the current methodologies developed for addressing out-of-distribution (OOD) issues in Section 3 and outline application scenarios in Section 4. We also discuss existing evaluation methods and available datasets before concluding with a discussion of the limitations of current methods and future directions.

% \section{Related Work}
% There is a line of research systematically analyzing the efforts for improving the generalization ability of NLP models. For example,
% \citet{schlegel2020beyond} provide an overview of the categorization of the weakness in models and datasets and the proposed methods to alleviate these weaknesses in NLI tasks.
% \citet{keith2020text} provide a review of text-related causal inference applications, methods for finding as well as evaluating causal relationships, and methods for removing confounding from text using causal estimates.
% \citet{feder2021causal} introduce the statistical challenge from the causal inference perspective to improve the performance and interpretability of NLP models. 
% \citet{koh2021wilds} propose a benchmark of 10 datasets to evaluate the data distribution shift mainly focusing on computer vision. 
% \citet{wang2021measure} present a unified survey of how to define, measure, and improve robustness in NLP.
% The most related to our work is \citet{arora2021types}, which is categorize examples into background shift and semantic shift and finds two major detection approaches (i.e., model calibration and density estimation). Different from existing work in specific task or technique, Our work tries to provide the first comprehensive review for analyzing the generalization challenge in NLP from an OOD perspective.

\begin{figure*}[t]
	\centering
	\includegraphics[width=\linewidth]{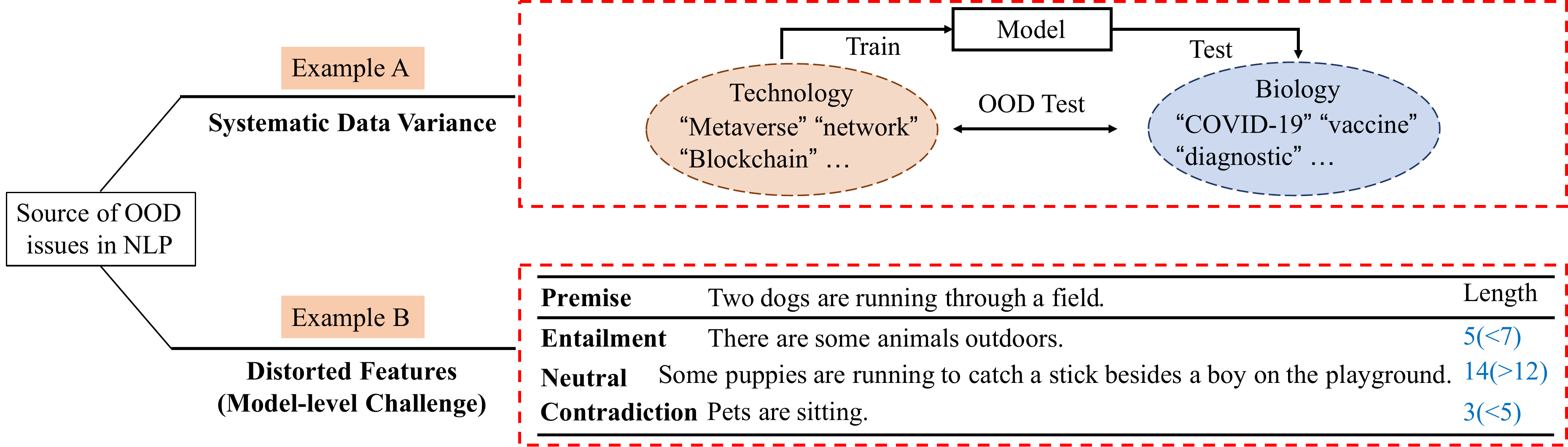}
	\caption{Taxonomy of OOD generalization scope and examples.}
	\label{fig:example}
\end{figure*}

\section{The Scope of OOD Generalization}\label{sec:scope}
Denote a set of labeled data as $ \mathcal{D} = \left\{ {\left( {{x_i},{y_i}} \right)} \right\}_{i = 1}^N$, where an input $x \in X$, output $y \in Y$, and $N$ is the number of datasets. A training dataset $\mathcal{D}_{train}=\{(X_{train},Y_{train})\}$ is generated by sampling from $\mathcal{D}$ with distribution $\mathcal{P}_{train}$, and the test dataset $\mathcal{D}_{test}=\{(X_{test},Y_{test})\}$ is sampled from $\mathcal{D}$ with distribution ${\mathcal{P}_{test}}$. \textbf{Out-of-distribution (OOD)} refers the circumstance when ${\mathcal{P}_{train}} \ne {\mathcal{P}_{test}}$. 

%In this section, we present the representative work in these two areas before discussing several consequences.

\subsection{Probabilistic view of OOD}
Let $X \in \mathcal{X}$ be the input space, $Y \in \mathcal{Y}$ be the output space, and $D$ be a training distribution defined on $\mathcal{X} \times \mathcal{Y}$. Suppose the true target distribution is $P_{\mathcal{X},\mathcal{Y}}$, which is close to but not identical to $D$ with $P_{\mathcal{X},\mathcal{Y}} \neq D$. An OOD sample is drawn from a distribution $Q_{\mathcal{X},\mathcal{Y}}$ which is significantly different from $D$, where $Q_{\mathcal{X},\mathcal{Y}}$ is assumed to have a support outside the support of $P_{\mathcal{X},\mathcal{Y}}$. A model $f: \mathcal{X} \to \mathcal{Y}$ is considered OOD if its performance on $Q_{\mathcal{X},\mathcal{Y}}$ is significantly worse than on $P_{\mathcal{X},\mathcal{Y}}$. The OOD detection function can be derived from a probabilistic perspective using Bayesian inference. In this case, we can estimate the posterior probability of a sample being OOD given its input features through Bayesian model averaging:
\begin{align*}
    P(\text{OOD} | x) = \sum_{\theta} P(\text{OOD}|\theta,x) P(\theta|x),
\end{align*}
where $\theta$ denotes the model parameters, $P(\text{OOD}|\theta,x)$ is the conditional OOD probability of model $f_\theta$ on input $x$, and $P(\theta|x)$ is the posterior probability of $\theta$ given $x$. The OOD detection function can be defined as a threshold on the posterior OOD probability:
\begin{align*}
    g(x) = [\max_y P(y|x) < \epsilon],
\end{align*}
where $\epsilon$ is a threshold parameter that determines the confidence of the prediction. The OOD detection performance can be evaluated using metrics such as the Receiver Operating Characteristic (ROC) curve or the Kolmogorov-Smirnov (KS) statistic.

\subsection{OOD in Text Classification}
Let $\mathcal{X}$ be the set of all possible documents, $\mathcal{Y}$ be the set of all possible labels, and $D$ be a training distribution defined on $\mathcal{X} \times \mathcal{Y}$. Suppose the true target distribution is $P_{\mathcal{X},\mathcal{Y}}$, which is close to but not identical to $D$ with $P_{\mathcal{X},\mathcal{Y}} \neq D$. An OOD sample is an input document drawn from a significantly different distribution $Q_{\mathcal{X}}$ where $Q_{\mathcal{X}}$ is assumed to have a vocabulary or language outside the support of $P_{\mathcal{X}}$. 

A text classification model $f: \mathcal{X} \to \mathcal{Y}$ is considered OOD if its performance on $Q_{\mathcal{X}}$ is significantly worse than on $P_{\mathcal{X}}$. The OOD detection function can be derived from a probabilistic perspective using Bayesian inference. In this case, we can estimate the posterior probability of a document being OOD given its bag-of-words features through Bayesian model averaging:

\begin{align*}
    P(\text{OOD} | \boldsymbol{x}) &= \sum_{\theta} P(\text{OOD}|\theta,\boldsymbol{x}) P(\theta|\boldsymbol{x})\\
    &= \sum_{\theta} \frac{P(\boldsymbol{x}|\text{OOD},\theta) P(\text{OOD}|\theta) P(\theta)}{P(\boldsymbol{x})},
\end{align*}

where $\theta$ denotes the model parameters, $\boldsymbol{x}$ is the bag-of-words representation of a document, $P(\text{OOD}|\theta)$ is the prior probability of the model being OOD assuming the model parameter $\theta$, $P(\boldsymbol{x}|\text{OOD},\theta)$ is the likelihood of observing the bag-of-words features $\boldsymbol{x}$ given that the document is OOD and the model parameter $\theta$, $P(\theta)$ is the prior probability of the model parameter $\theta$, and $P(\boldsymbol{x})$ is the marginal likelihood of observing the bag-of-words features $\boldsymbol{x}$. 

The conditional OOD probability of model $f_\theta$ on input document $\boldsymbol{x}$ given the parameter $\theta$ is defined as:
\begin{align*}
    P(\text{OOD}|\theta,\boldsymbol{x}) = \frac{P(\boldsymbol{x}|\text{OOD},\theta) P(\text{OOD}|\theta)}{P(\boldsymbol{x}|\theta)}.
\end{align*}

The OOD detection function can be defined as a threshold on the posterior OOD probability:

\begin{align*}
    g(\boldsymbol{x}) = [\max_{y} P(y|\boldsymbol{x})< \epsilon],
\end{align*}

where $P(y|\boldsymbol{x})$ is the posterior probability of the document belonging to class $y$ given the bag-of-words features $\boldsymbol{x}$, and $\epsilon$ is a threshold parameter that determines the confidence of the prediction.

The OOD detection performance can be evaluated using metrics such as the Receiver Operating Characteristic (ROC) curve or the Kolmogorov-Smirnov (KS) statistic, which capture the trade-off between true positive rate and false positive rate, or the maximum distance between the cumulative distribution functions of the OOD and in-distribution predictions, respectively.
% systematic variance (Sec.\ref{sec:systematic_var}), variation in expressions (Sec.\ref{sec:flex}), and training data issues (Sec.\ref{sec:data_artifact}).

\section{Taxonomy of Out-of-Distribution (OOD) Problems}

In the field of text classification, OOD problems can be classified based on their source, which includes data-level variance and model-level bias during the learning process. To organize the existing NLP research on OOD problems, we classify them into two main areas, as depicted in Figure \ref{fig:example}, namely \textit{Systematic Data Variance} and \textit{Distorted Features}. While systematic data variance encompasses the domain generalization problem, "distorted features" represent a range of issues typically caused by shortcut learning resulting from inductive reasoning. 

\subsection{Systematic Data Variance}\label{sec:systematic_var}
Addressing the problem of systematic data variance involves utilizing domain generalization methods, assuming the unavailability of labeled or unlabeled data from the target domain. Previous studies have explored this approach in sentiment analysis  (SA)~\cite{kaushik2019learning,ni2019justifying,yang2021exploring,lu2022rationale}, natural language inference (NLI)~\cite{williams2018broad,wang2018glue,Varshney2020TowardsIS,hendrycks2020pretrained}, and named entity recognition (NER)~\cite{jia2020multi,plank2021cross}. Different domains have intrinsically different feature distributions, and instances from different domains have different predicted vocabulary distributions, which leads to the OOD generalization challenge to some extent as shown in Figure \ref{fig:example} (Example A). 

Numerous NLP studies have aimed to tackle systematic variations between training and testing distributions, encompassing a vast body of literature on domain generalization \cite{blitzer2006domain,pan2010cross,ganin2016domain,zhu2009introduction,ruder2018strong,han2019unsupervised,guo2020multi,lim2020semi} and cross-task transfer \cite{Johnson2017GooglesMN,wang2022miner,levy2017zero,eriguchi2018zero}, as shown in Table~\ref{table:scope_sys}. These studies can be broadly categorized into input-level variation and output-level variation. Notable comprehensive surveys dedicated to this task include that by \citet {ramponi2020neural} and \citet{wang2021measure}.

%\footnote{Considering our survey spans too many tasks in NLP, We only summarize the a portion of representative work in this table (the same as the following tables).}

\begin{table*}[t]
\centering
\setlength{\belowcaptionskip}{-0.5cm} 
\resizebox{\linewidth}{!}{
\begin{tabular}{@{}c|c|c|c|c@{}}
\toprule
\textbf{Data Variance} & Task & Papers & Key Methods & Typical Datasets \\ \midrule
& \begin{tabular}[c]{@{}c@{}}Sentiment\\Analysis\end{tabular} &
\parbox[c]{7cm}{
\citet{ganin2016domain,chen2018multinomial,laparra2020rethinking}.} & \parbox[c]{5cm}{Adversarial learning ; \\ Multinomial adversarial networks.}   & \parbox[c]{4cm}{Amazon Reviews; \\ IMDB Reviews.} \\  \cmidrule(l){2-5} 
& MT  & \parbox[c]{7cm}{\citet{belinkov2017synthetic,khayrallah-koehn-2018-impact}.} &  \parbox[c]{5cm}{Invariant representation learning; \\ Training on adversarial examples.}   & \parbox[c]{4cm}{
%NATURAL NOISE: 
WiCoPaCo; RWSE Wikipedia; \\ Revision Dataset; \\ MERLIN corpus.

%SYNTHETIC NOISE
%Apply Swap, Middle Random, Fully Random, and Keyboard Typo to the four dataset mentioned above

%WMT 2017\tablefootnote{Add noise to WMT 2017}
}    \\  \cmidrule(l){2-5} 
& \begin{tabular}[c]{@{}c@{}}Label-sharing\\NER\end{tabular} & \parbox[c]{7cm}{\citet{liu2021noisy,huang2021nsrl}.} &  \parbox[c]{5cm}{\leftline{Noisy supervised pre-training;} Calibrated confidence methods.}    & \parbox[c]{4cm}{CoNLL2003; Tweet; Webpage; Wikigold.}      \\ \cmidrule(l){2-5}
\multirow{10}{*}{\begin{tabular}[c]{@{}c@{}}Annotation \\artifacts\end{tabular}}           & QA     &  \parbox[c]{7cm}{
\citet{cai2017pay,min2019compositional,bartolo2021models,bartolo2021improving,lyu2022extending}.}  & \parbox[c]{5cm}{

Generator-in-the-loop models.

}    & \parbox[c]{4cm}{\leftline{ROC Story; HOTPOTQA;} NewsQA; SQuAD1.1;\\ AdversarialQA.}  \\  \cmidrule(l){2-5} 
& NLI  & \parbox[c]{7cm}{\citet{poliak2018hypothesis,naik2018stress,zellers-etal-2018-swag,feng2019misleading,mccoy2019right,le2020adversarial,sakaguchi2020winogrande,nie-etal-2020-adversarial,liu-etal-2020-hyponli,gardner2021competency,pezeshkpour2022combining,wu2022generating}.}  & \parbox[c]{5cm}{Data augmentation; \\ Human-and-model-in-the-loop; \\ Adversarial filtering; \\ Training on adversarial examples.}   & \parbox[c]{4cm}{Stress Test; ANLI; SWAG; HANS; SNLI-hard; MultiNLI-hard.
}   \\ \cmidrule(l){2-5} 

& \begin{tabular}[c]{@{}c@{}}MRC \end{tabular}  & \parbox[c]{7cm}{ \citet{kaushik2018much,sugawara2018makes,sugawara2020assessing,bartolo2020beat,lai2021machine}.}  & \parbox[c]{5cm}{
Shortcut investigation.}  & \parbox[c]{4cm}{\leftline{bAbI; SQuAD; CBT; CNN;} Whodid-What; DuoRc.

%How Much Reading Does Reading Comprehension Require? A Critical Investigation of Popular Benchmarks
%compare  bAbI, SQuAD, CBT, CNN, and Who- did-What datasets. make a conclusion that SQuAD and CNN appear better-constructed.

}\\ \cmidrule(l){2-5} 
& MT &\parbox[c]{7cm}{\citet{vanmassenhove2018getting,stanovsky2019evaluating,tomalin2021practical,choubey2021improving}.} &\parbox[c]{5cm}{Adversarial learning; \\ Gender-filtered self-training.}    & \parbox[c]{4cm}{WinoMT; MuST-SHE.}   \\ \cmidrule(l){2-5} 
& \begin{tabular}[c]{@{}c@{}}Coreference\\ Resolution \end{tabular}   & \parbox[c]{7cm}{\citet{rudinger-etal-2018-gender,zhao2018gender}.}  & \parbox[c]{5cm}{Data augmentation.}    & \parbox[c]{4cm}{WinoBias; \\ Winogender Schemas.}    \\ \cmidrule(l){2-5} 
& Toxicity Detection & \parbox[c]{7cm}{\citet{park2018reducing,Dixon2018MeasuringAM}.}  & \parbox[c]{5cm}{
%Reducing Gender Bias in Abusive Language Detection
Data augmentation;\\ Debias word embeddings.
}  & \parbox[c]{4cm}{
%Reducing Gender Bias in Abusive Language Detection
%Word Embedding Association Test (WEAT) % measure model bias inside word embeddings
%SemEval % evaluating racial and gender bias of those sys- tems
%Sexist Tweets (st);\\ Abusive Tweets (abt)

%Measuring and Mitigating Unintended Bias in Text Classification
%Wikipedia Talk Pages
%I think those datasets are importants
Sexist Tweets (st);\\ Abusive Tweets (abt).
} 
     \\ \midrule

\multirow{2}{*}{Output variance}
    & \begin{tabular}[c]{@{}c@{}}Label-Different\\ NER  \end{tabular} & \parbox[c]{7cm}{
\citet{snell2017prototypical,ghaddar2017winer,wu2020single,Nguyen2021DOZENCZ,cui2021template,ma2021template,zhou2021melm,lee2021good,das2021container,wang2022miner}.}  & \parbox[c]{5cm}{Self-training methods; Prompt-based methods; Information theories; Prototype-based methods; Distance-based methods; Knowledge-enhanced methods.}  &\parbox[c]{4cm}{\leftline{CoNLL2003; MIT Movie;} MIT Restaurant; WNUT2017; Ontonotes 5.0 Dataset; BioNER.} \\ \cmidrule(l){2-5} 
& \begin{tabular}[c]{@{}c@{}}Machine \\ Translation\end{tabular}   & \parbox[c]{7cm}{\citet{Johnson2017GooglesMN,zhang2020improving,arivazhagan2019missing,ji2020cross,liu2021improving}.} & \parbox[c]{5cm}{Multilingual corpus pre-training; \\Back translation; \\Invariance representation learning; \\Language independent representations learning.}  &\parbox[c]{4cm}{
WMT'14; WMT'17; Newstest 2012; Newstest 2013; Newstest 2016; Newstest 2015; IWSLT 2017.}       \\ \bottomrule
\end{tabular}
}
\caption{OOD generalization challenges related to the systematic data variance.}
\label{table:scope_sys}
\end{table*}
%  [CITE SURVEYS ON domain generalization, TRANSFER LEARNING IN NLP].}

However, another line of research focuses on addressing distributional variances caused by issues in the quality of training data. This type of variation also encompasses the OOD problem from a data perspective but is typically attributed to data noise, including annotation mistakes \cite{belinkov2017synthetic,rychalska2019models,moradi2021evaluating,naplava2021understanding} and unintended biased distributions \cite{zhao2019gender,maudslay2019s,zhong2020does,gardner2020evaluating,rudinger2017social}. Previous studies have highlighted the significant impact of data noise on NER models, with notable examples being the CoNLL2003 dataset \cite{sang2003introduction}. Such mistakes can stem from various sources, such as pseudo, weak, or distant annotations \cite{liu2021noisy}.

Furthermore, OOD generalization tasks involving variance in label types have been extensively investigated in NLP across various tasks. Examples include constituent parsing \cite{tateisi-etal-2005-syntax,fried-etal-2019-cross,yang-etal-2022-challenges}, NER \cite{ghaddar2017winer,cui2021template,das2021container,lee2021good}, machine translation \cite{Johnson2017GooglesMN,zhang2020improving,arivazhagan2019missing,ji2020cross,liu2021improving}, and QA \cite{lee-etal-2019-domain}. For instance, in constituent parsing, the distribution of output constituent labels may differ between news and biomedicine domains \cite{fried-etal-2019-cross}. In NER, named entity labels can vary significantly across different domains, such as CHEMICALs and SYMPTOMs in the biomedical domain \cite{tarcar2019healthcare}, or PERSONs and LOCATIONs in news articles \cite{lee2021good}. Variations in label sets can also arise in NLI tasks \cite{mccoy2019right} and generalization between unseen datasets in QA \cite{lee-etal-2019-domain,niu2023learning}.

\begin{table*}[t]
\centering
\setlength{\belowcaptionskip}{-0.5cm} 
\resizebox{\linewidth}{!}{
\begin{tabular}{@{}c|c|c|c|c@{}}
\toprule
\multicolumn{1}{l|}{\begin{tabular}[c]{@{}c@{}}\textbf{Flexibility of}\\ \textbf{Expression}\end{tabular}}                                 & Task               & Papers      & Key Methods        & Typical Datasets                                                                                          \\ \midrule
                                                          
& Text Classification         & \parbox[c]{6cm}{\citet{oren2019distributionally,hendrycks2020pretrained,wang2021identifying,du2021towards,liu2021just,moradi2021evaluating,naplava2021understanding,wang-etal-2021-textflint}.}   &  \parbox[c]{6cm}{Data augmentation; \\ Regularization on shortcuts; \\ Spurious features identification \& removal; \\Distributionally robust optimization (DRO).}  & \parbox[c]{4cm}{WildNLP; TextFlint; \\IMDB Reviews; \\ Kindle Reviews.      %\\ GLUE
}    \\  \cmidrule(l){2-5}                       
& \begin{tabular}[c]{@{}c@{}}Natural Language\\ Generation\end{tabular} & \parbox[c]{6cm}{\citet{cheng-etal-2019-robust,zhang2019bridging,zhou2021distributionally,hewitt2021ensembles}.} &  \parbox[c]{6cm}{
%Robust Neural Machine Translation with Doubly Adversarial Inputs
Adversarial attack learning; 
%Bridging the Gap between Training and Inference for Neural Machine Translation
%hard to categorize the method
%Distributionally Robust Multilingual Machine Translation
\\Group DRO; 
%Ensembles and Cocktails: Robust Finetuning for Natural Language Generation
Robust fine-tuning.
}  & \parbox[c]{4cm}{\leftline{NIST; WMT'14; WevNLG;} XSUM; Open-domain QA.} \\ \cmidrule(l){2-5}   
\multirow{7}{*}{\begin{tabular}[c]{@{}c@{}}Compositional\\ generalization\end{tabular}}
& Evaluations                 & \parbox[c]{6cm}{\citet{czarnowska2019don,kaushik2019learning,gardner2020evaluating,warstadt2020blimp,hu2020xtreme,lewis2020mlqa,lazaridou2021mind,liu2021challenges,koh2021wilds,chen2022can}.} &          \parbox[c]{6cm}{Contrast sets; Fine-grained evaluations.}   & \parbox[c]{4cm}{
%Don’t Forget the Long Tail! A Comprehensive Analysis of Morphological Generalization in Bilingual Lexicon Induction
%Bilingual Lexicon Induction任务不典型 不需要列出来

%LEARNING THE DIFFERENCE THAT MAKES A DIFFER- ENCE WITH COUNTERFACTUALLY-AUGMENTED DATA
%IMDb; SNLI %(counterfactual)

%Evaluating Models’ Local Decision Boundaries via Contrast Sets
%DROP; IMDb

%BLiMP: The Benchmark of Linguistic Minimal Pairs for English
%BLiMP 小众任务 不讨论

%XTREME: A Massively Multilingual Multi-task Benchmark for Evaluating Cross-lingual Generalization
%XTREME; % a multi-task benchmark for evaluating the cross-lingual generalization capabilities

%MLQA: Evaluating Cross-lingual Extractive Question Answering
%MLQA

%Mind the Gap: Assessing Temporal Generalization in Neural Language Models
%WMT; CUSTOMNEWS; ARXIV

%Challenges in Generalization in Open Domain Question Answering
%Open Natural Question; TriviaQA; WebQuestions 

%Wilds: A benchmark of in-the-wild distribution shifts.
%Wilds ->Amazon and Yelp reviews%mention一下

%Can Rationalization Improve Robustness?
%beer review; FEVER; SQuAD

%我选择
% SQuAD; WMT; IMDb.
\leftline{BLiMP; XTREME; MLQA;} ARXIV; Wilds; SQuAD.

}      \\  \cmidrule(l){2-5} 
& NLU                         & \parbox[c]{6cm}{\citet{lake2018generalization,russin2019compositional,li2019compositional,gordon2019permutation,andreas2020good,keysers2020measuring,kim2020cogs,kim2021improving}.}          & \parbox[c]{6cm}{ Dedicated train objects;\\ Structure annotation.
}       & \parbox[c]{4cm}{

SCAN; CFQ; COGS.
}          \\ \cmidrule(l){2-5}   
& Semantic Parsing            & \parbox[c]{6cm}{\citet{iyer2017learning,lake2018generalization,dong2018coarse,lake2019compositional,yu2019sparc,furrer2020compositional,kim2021sequence,gupta2022structurally}.}    &  \parbox[c]{6cm}{Span-level supervised attention; \\ Human-in-the-loop; \\ Meta sequence-to-sequence learning; \\ Structurally diverse sampling.}  & \parbox[c]{4cm}{
%Learning a Neural Semantic Parser from User Feedback
%GEO880 \\ ATIS \\ SCHOLAR

%Generalization without Systematicity: On the Compositional Skills of Sequence-to-Sequence Recurrent Networks
%SCAN %SCAN (unlike the original CommAI tasks) can be straightforwardly treated as a supervised sequence- to-sequence semantic parsing task 

%Coarse-to-Fine Decoding for Neural Semantic Parsing
%GEO \\ ATIS \\ DJANGO \\ WIKISQL

%Compositional generalization through meta sequence-to-sequence learning
%SCAN

%Sparc: Cross- domain semantic parsing in context.
%SParC %a dataset for cross-domain Semantic Parsing in Context
%ATIS, GeoQuery, SCONE, SequentialQA
%WikiSQL \\ Spider %other cross-domain semantic parsing datasets

%Compositional Generalization in Semantic Parsing: Pre-training vs. Specialized Architectures
%SCAN \\ CFQ % Compositional Freebase Questions (CFQ)

%Sequence-to-sequence learning with 1123 latent neural grammars.
%SCAN

%Structurally Diverse Sampling Reduces Spurious Correlations in Semantic Parsing Datasets
%COVR \\ ATIS \\ Overnight \\ Schema2QA \\ SM-CalFlow

%I this those datasets are important
ATIS; GEO; SCAN; CFQ.
}  \\ \cmidrule(l){2-5}   
& \begin{tabular}[c]{@{}c@{}}Machine\\ Translation\end{tabular} & \parbox[c]{6cm}{\citet{chen2020compositional,li2021compositional,zheng2021disentangled}.}   &  \parbox[c]{6cm}{Neural symbolic stack machines; \\ Representation disentanglement.}   & \parbox[c]{4cm}{

CoGnition; SCAN. % the datasets after is mainly for compositional generalization\\ SCAN  \\ CFQ \\ COGS
}  \\  \cmidrule(l){2-5}
& QA & \parbox[c]{6cm}{\citet{gu2021beyond,lewis2021question,bogin2021latent}.}  & \parbox[c]{6cm}{Data augmentation; Prompt-tuning; \\Continual pre-training.}   & \parbox[c]{4cm}{%KBQA \\ GRAILQA \\%Beyond I.I.D.: Three Levels of Generalization for Question Answering on Knowledge Bases
%\\ SIMPLEQ \\ GRAILQA \\ WEBQ \\COMPLEXWEBQ \\ GRAPHQ

%Question and Answer Test-Train Overlap in Open-Domain Question Answering Datasets
%WebQuestions \\ TriviaQA \\ Open Natural Questions \\

% Latent Compositional Representations Improve Systematic Generalization in Grounded Question Answering
%CLOSURE \\ CLEVR \\ 

% very typical, but is not being tested in papers
%SQuAD

%I would choose 
GRAILQA; TriviaQA; \\Open Natural Questions;  \\WebQuestions.}  \\ \midrule
Logic reasoning  & MRC  & \parbox[c]{6cm}{\citet{dong2016language,yu2019reclor,rogers2021qa,liu2021logiqa,zhong2021ar,huang2021dagn}.}   & \parbox[c]{6cm}{GAN; Graph neural networks;\\ Knowledge-enhanced methods.}   & \parbox[c]{4cm}{

%Language to Logical Form with Neural Attention
%JOBS \\ GEO \\ ATIS \\ IFTTT \\

%RECLOR: A READING COMPREHENSION DATASET REQUIRING LOGICAL REASONING
%RECLOR \\ DREAM \\ MCTest \\ ARC \\ RACE

%LogiQA: A Challenge Dataset for Machine Reading Comprehension with Logical Reasoning
%LogiQA \\ COSMOS \\ Social IQa

%AR-LSAT: Investigating Analytical Reasoning of Text
%AR-LSAT

%Dagn: Discourse-aware graph 1020 network for logical reasoning
% typical
%SQuAD \\ CLEF QA \\ NewsQA \\ RACE \\ BoolQ \\ MS MARCO \\ CoQA \\ COSMOS

%I would choose 
SQuAD; DROP; LogiQA; \\ HotpotQA;  ReClor; AR-LAST.

}   \\  \cmidrule(l){2-5}   
&\begin{tabular}[c]{@{}c@{}} Mathematical \\Problem \end{tabular}  & \parbox[c]{6cm}{\citet{brown2020language,cobbe2021training,drori2021neural,hendrycks2021measuring}}    & \parbox[c]{6cm}{Self-supervised training (GPT3); \\ Training verifiers; Program synthesis (Codex).}  & \parbox[c]{4cm}{
%Training Verifiers to Solve Math Word Problems
%GSM8K \\ Dolphin18K \\ AQuA-RAT \\ MathQA \\ Ape210K \\ ASDiv \\ The MATH dataset

%Language Models are Few-Shot Learners
%To test GPT-3’s ability to perform simple arithmetic operations without task-specific training, we developed a small battery of 10 tests that involve asking GPT-3 a simple arithmetic problem in natural language:

%April 2022: A Neural Network Solves, Explains, and Generates University Math Problems by Program Synthesis and Few-Shot Learning at Human Level
%MATH

%Measuring Mathematical Problem Solving With the MATH Dataset
%MATH
%pretrained on Auxiliary Mathematics Problems and Solutions (AMPS)
%DeepMind Mathematics dataset

MATH Datasets; \\\leftline{DeepMind Datasets.}
}  \\  \bottomrule
\end{tabular}
}
\caption{OOD generalization challenges related to distorted features learned by models.}
\label{table:scope_flex}
\end{table*} 

% \noindent\textbf{OOD Detection}
% It is noteworthy that although the OOD detection \cite{hendrycks2017baseline,hendrycks2020pretrained,fort2021exploring} has surfaced in this domain, it receives relatively less attention than developing methods for improving the OOD robustness. Meanwhile, the fine-grained diagnostics of model behavior are rarely found in evaluations of the OOD robustness. Based on the fact that we still lack a standard definition of OOD examples and fine-grained evaluations, the efficacy of methodological papers on improving OOD robustness may need to be carefully validated.

% For instance, for robustness research, we expect a model to give consistent outputs regardless of expressional variances such as paraphrasing and change of punctuation.
% For logical and commonsense reasoning, the underlying solution does not necessarily depends on surface-level text distributions, and a model is expected to use logical inference or external knowledge for addressing OOD issues in surface text.

\subsection{Distorted Features}\label{sec:flex}
Models' predictions are often influenced by distorted features learned from spurious patterns between training data and labels, as well as existing shortcuts in the dataset. For instance, as illustrated in Example B, sentence length has inadvertently become a learned feature during training, where 60\% of the heuristics in entailment examples have 7 or fewer tokens, and half of the hypotheses with more than 12 or less than 5 tokens are neutral or entailment, respectively \cite{gururangan2018annotation}. In the field of NLP, there is a strand of research on OOD generalization that addresses the discrepancy between the \textit{internal} distribution of abstract tasks and the \textit{external} distribution of language expressions. These problems involve compositional generalizations and logical reasoning, where models often struggle to adapt to the external distributions, as summarized in Table~\ref{table:scope_flex}.

\textbf{Compositional generalization} refers to the challenge of learning the distribution of atoms given the surface distributions of their compositions. It has garnered significant attention in NLP research, encompassing areas such as semantic parsing \cite{iyer2017learning,gupta2022structurally}, QA \cite{gu2021beyond,lewis2021question}, machine translation \cite{li2021compositional}, and general natural language understanding (NLU) tasks \cite{lake2018generalization,keysers2020measuring}. Researchers \cite{keysers2020measuring,kim2021improving} have found that state-of-the-art neural models struggle to generalize to novel compounds in a manner similar to human performance. Several benchmarks have been introduced to evaluate compositional generalization. For example, the SCAN dataset by \citet{lake2018generalization} is designed for sequence-to-sequence generalization \cite{russin2019compositional,li2019compositional,gordon2019permutation,andreas2020good}. Additionally, \citet{keysers2020measuring} and \citet{kim2020cogs} propose the CFQ and COGS benchmarks, respectively, for semantic parsing. \citet{li2021compositional} propose the CoGnition dataset to assess how neural machine translation models generalize to novel compounds \cite{hupkes2020compositionality,zheng2021disentangled,dankers2021paradox,jung2022machine}.

To address the challenges of compositional generalization, achieving \textbf{OOD robustness} is highly desirable as current NLP models have shown fragility to variations in expression, where even minor punctuation changes can lead to different outputs \cite{wang-etal-2021-textflint}. Furthermore, \citet{moradi2021deep} observe significant performance decay of NLP models in domain-specific tasks, such as the clinical domain, due to noise, grammar errors, missing words, punctuation, typos, and other factors. Additionally, \citet{wang-etal-2021-textflint} develop a unified multilingual robustness evaluation platform for NLP tasks to provide comprehensive robustness analysis. In formal terms, OOD robustness typically refers to the model's ability to maintain reliable predictions on test data with unintentional distribution shifts, while avoiding the interference of shortcuts or spurious features. For example, in the following example, the addition or removal of punctuation can alter the relationship between the hypothesis and premise. The hypothesis is \textit{``I thank my mother, Anna, Smith and John.''}, and the premise is \textit{``I thank four people.''} The hypothesis is true (entailment) given the premise. However, if we remove a comma, the hypothesis becomes \textit{``I thank my mother Anna, Smith and John.''}, which changes the relationship to a contradiction. %Such robustness issues have also been investigated for other tasks, including sentiment analysis~\cite{Zhang2021InterpretingTR}, NLI~\cite{ek2020does}, and parsing~\cite{sogaard2018nightmare}. 

A strand of work on OOD robustness considers whether a model learns the most relevant features \citep{xing2020tasty,hendrycks2020pretrained,wang-etal-2021-textflint,gokhale2022generalized}. In particular, it has been shown that neural models can learn ``\textbf{spurious features}'' \cite{tu2020empirical}, or statistical features that represent the surface statistical distribution existing in data distribution of the certain domain instead of the decisive causal relation, such as position cues in machine reading comprehension \cite{kaushik2018much}, hypothesis-only bias \cite{poliak2018hypothesis}, syntax heuristics \cite{mccoy2019right} in NLI, superficial cues in common sense reasoning \cite{kavumba2019choosing}, and spurious patterns in sentiment analysis \cite{wang2021robustness}). For example, in ``Jonny’s movies are great.'', the decisive feature for the ``positive’’ sentiment is ``great’’, and ``Jonny'' is a spurious feature. The spurious feature ``Jonny'' can cause a model to incorrectly give a ``negative’’ sentiment output for the input sentence ``Cathy’s movies are great.''. 

Another angle of OOD robustness is to design the \textbf{challenging dataset}. For example, the recently proposed contrast sets~\cite{kaushik2019learning,gardner2020evaluating,warstadt2020blimp} reveal the failure of capturing true underlying distributions, which show the fragility of models against small variations of input expressions. In addition, although researchers also propose the benchmark to reveal the importance of OOD detection~\cite{hendrycks2017baseline,hendrycks2020pretrained,fort2021exploring}, there is a consensus that the community still lacks a standard definition of OOD examples and fine-grained evaluations. 

\section{Application Scenarios}
We introduce two representative application scenarios affected by the OOD generalization as follows. 

\subsection{Deployment in High-state Domains} Current NLP methods tend to learn implicitly superficial cues instead of the causal associations between the input and labels, as evidenced by~\cite{geirhos2020shortcut}, and thus usually show its brittleness when deploying in the real-world scenario. Despite the generalization ability of LLMs, such as ChatGPT has achieved great process. Still, the relatively low generalization ability of medium-size models hinders the deployment of NLP systems, especially for high-state domains, from health and medicine to finance and business, and should be taken more seriously. Recent work \cite{sugawara2018makes,sugawara2020assessing,lai2021machine,wang2021identifying,du2021towards,zhu2021quantifying,bastings2021will} indicates that current PLMs unintentionally learn shortcuts to trick specific benchmarks and such tricks (i.e., syntax heuristics, lexical overlap, and relevant words) that use partial evidence to produce unreliable output which is in the open domain. Notably, a recent comprehensive evaluation of OOD generalization in text classification \cite{yang2022glue} shows that the average accuracy of PLMs on cross-domain evaluations falls significantly short of human performance, even for the highest-performing model (\textbf{80.1\% -- human versus 74.6\% -- model}). In contrast to GLUE, where over 20 single-model results outperform human baselines, none of the backbones included in OOD tests is able to surpass human performance. Lacking the enough OOD generalization ability is also related to social bias. 

\subsection{Social Bias} Recent studies \cite{gardner2020evaluating} have uncovered a problematic tendency for gender bias in sentiment analysis \cite{zmigrod2019counterfactual,maudslay2019s,lu2020gender}. Bias exists in different forms of language representations, including word embeddings \cite{bolukbasi2016man,caliskan2017semantics,zhao2018learning,gonen2019lipstick}, contextualized word embeddings \cite{zhao2019gender} and sentence embeddings \cite{may2019measuring}. Some found that the embeddings of femininity words and masculinity words are often clustering into different groups (e.g., occupation)~\cite{gonen2019lipstick,bolukbasi2016man,caliskan2017semantics,zhao2018learning,zhao2019gender,gonen2019lipstick}.
Gender bias also affects the coreference resolution system, which tends to link a pronoun to occupations dominated by the pronoun gender~\cite{rudinger-etal-2018-gender,zhao2018gender}. In machine translation, \citet{vanmassenhove2018getting} and \citet{stanovsky2019evaluating} find that models tend to make stereotypical assignments of gender roles when translating occupation words. Apart from gender bias, there are other forms of social bias in NLP data, such as disability \cite{hutchinson-etal-2020-social}, race \cite{kiritchenko2018examining}, age \cite{agebias}, etc.

% Learning spurious features can cause a model to demonstrate weak robustness. In the example above, the spurious feature ``Honey’’ can cause a model to incorrectly give a ``negative’’ sentiment output for the input sentence ``Cathy’s movies are great.’’. 

\begin{figure}[t]
	\centering
	\setlength{\belowcaptionskip}{-0.5cm} 
	\includegraphics[width=\linewidth]{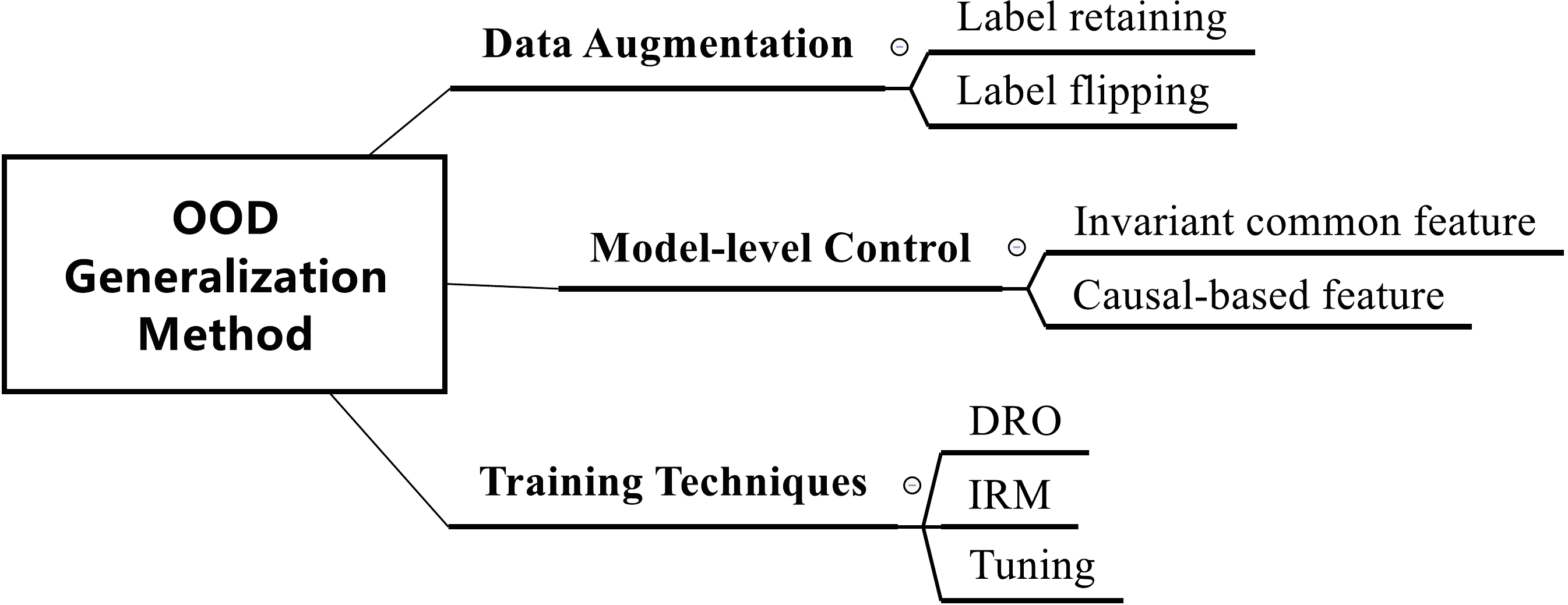}
	\caption{Improving the OOD generalization performance towards text classification.}
	\label{fig:method}
\end{figure}

\section{Methods}\label{sec:method}
Existing work to address OOD issues in NLP can be categorized into three groups: data augmentation~(Sec.\ref{sec:da}), model-level control (Sec.\ref{sec:model_control}), and training techniques (Sec.\ref{sec:train_tech}) shown in Fig.~\ref{fig:method}. Details of methods towards OOD generalization according to different tasks are provided in Table~\ref{table:full_1}, Table~\ref{table:full_2}, and Table~\ref{table:full_3} in Appendix.

\subsection{Data Augmentation}\label{sec:da}
Data Augmentation (DA) refers to strategies for increasing the diversity of training data without explicitly collecting new data \cite{feng2021survey}, which is beneficial to the generalization of NLP models by reducing the overfitting and improving the robustness. 
Existing surveys \cite{hedderich2021survey,feng2021survey,bayer2021survey,chen2021empirical,li2022data} discuss data augmentation in low-resource NLP scenarios from different perspectives, and \citet{chen2021empirical} provide an empirical study over different DA methods. 
Here we focus on DA for OOD generalization.

% \subsubsection{Label Retaining}
\noindent\textbf{Semi-fact.} A typical type of augmentation data method in NLP is to substitute part of the content or add perturbation in original data, which mainly focuses on improving the diversity of original data and not changing the semantic meaning or label. \citet{zhang2015character}, \citet{miao2020snippext} and \citet{yue2020phicon} substitute words or entities using synonyms. Perturbation implementation in the original data usually consists of tokens in the sentences~\cite{zhang2018mixup,wei2019eda,miao2020snippext,xie2020unsupervised,zhao2019gender,zhao2018gender}. \citet{Miyato2017AdversarialTM,cheng-etal-2019-robust,zhu2019freelb,jiang2020smart,zheng2020out} improve the generalization and robustness by adding adversarial perturbations to the original data. \citet{kumar2020data} and \citet{lu2022rationale} adopt large pre-trained models (GPT-2~\cite{Radford2019LanguageMA}, BART~\cite{liu2019roberta}, BERT~\cite{devlin2018bert}) to generate conditional data augmentation. \citet{lu2022rationale} propose a retionale-centre framework with human-in-the-loop by using semi-factual augmentations.
% and human-intervened corrections to decouple spurious associations and bias model.

% \citet{lu2022rationale} propose a semi-factual augmentations method to  mitigate spurious associations via human-in-the-loop.

% \citet{lu2022rationale} propose a retionale-centre framework with human-in-the-loop by using semi-factual augmentations and human-intervened corrections to decouple spurious associations and bias model.  

% The semi-factual example is generated by replacing some non-rationales with synonyms from BART.

% \subsubsection{Label Flipping}
\noindent\textbf{Counterfactual} data augmentation (CDA) is widely adopted to mitigate bias in neural NLP tasks by operating on biased text \cite{maudslay2019s,zmigrod2019counterfactual}. A counterfactual example constructed by flipping the label helps to learn real associations between input and label. For instance, \citet{lu2020gender} proposes CDA method to mitigate gender bias in neural coreference resolution, which is a generic methodology for corpus augmentation via causal interventions (i.e., breaking associations between gendered and gender-neutral words). 
In sentiment analysis, \citet{kaushik2019learning}, \citet{kaushik2020explaining}, and \citet{Wang2020IdentifyingSC} employ humans for generating counterfactual data and has shown to be effective to mitigate the influence of spurious patterns. Automatically counterfactual generation has been also well-studied \cite{yang2021exploring,wang2021robustness,wu2021polyjuice}.
% \looseness=-1

% Researchers have found that artifacts threaten the model's validity and reliability.  

\subsection{Model-level Control}\label{sec:model_control}
% \subsubsection{Invariant Common Feature}
Feature representation learning is one of the keys to OOD generalization. We review model-level methods from the perspectives of the invariance and causal factor.

\noindent\textbf{Invariant Common Feature} 
There has been a long-lasting line of research investigating invariant features for transfer learning \cite{blitzer2006domain,Johnson2017GooglesMN,li2019transferable,du2020adversarial,wang2022miner}. For discrete linear models, structured correspondence learning makes use of unlabeled target-domain data \cite{blitzer2006domain}, \citet{daume2009frustratingly} leverages labeled data, and \cite{johnson2005high} uses unlabeled data For neural models, a commonly used technique is adversarial learning \cite{Goodfellow2015ExplainingAH,ganin2016domain,zhang2019interactive}, where a domain classifier is trained with adversarial loss to drive out domain-specific information in a hidden layer representation for cross-domain \cite{liu2018learning,li2019transferable,du2020adversarial} or cross-task decision making \cite{Johnson2017GooglesMN,levy2017zero,eriguchi2018zero,lee-etal-2019-domain,wang-etal-2019-adversarial,keung-etal-2019-adversarial,vernikos2020domain,wang2022miner}.

Zero-shot methods are also adopted to learn invariant features, which requires OOD generalization on unseen tasks. For instance, in machine translation, \citet{Johnson2017GooglesMN}, \citet{arivazhagan2019missing},        \citet{ji2020cross}, and \citet{liu2021improving} train a translation model to learn language independent representation, which helps the model generalize to unseen language pairs.
In sentiment analysis, \citet{liu2018learning} and \citet{du2020adversarial} conduct the adversarial training to derive the enhanced domain-invariant features for cross-domain classification. 

% One line of OOD generalization problem refers to the domain generalization. \rx{\sout{The trained model in the source domain is expected to generalize well to the target domain at the feature level.}} Several typical transductive methods learn invariant features directly. \citet{blitzer2006domain} propose a structural correspondence learning (SCL) to make use of the unlabeled data from the target domain to extract common features that reduce the performance difference between domains.
% \citet{daume2009frustratingly} propose a kernel-mapping function for NLP problem, which maps the data from both source and target domains to a high-dimensional feature space, where standard discriminative learning methods are used to train the classifier. In addition, to extract domain-invariant features, adversarial learning~\cite{Goodfellow2015ExplainingAH,ganin2016domain,zhang2019interactive} methods are proposed to train the feature generator to minimize the text classification and simultaneously deceive the discriminator~\cite{liu2018learning,li2019transferable,du2020adversarial,vernikos2020domain}.

% \subsubsection{Causal-based Feature}
\noindent\textbf{Causal-based Feature} Causal inference aims to find the effectiveness of one variable on another variable \cite{Holland1986StatisticsAC,morgan2015counterfactuals,Imbens2015CausalIF,pearl2000models}. Because the relationships between the causal features and the labels are invariant under distribution shift~\cite{pearl2000models,quionero2009dataset}, learning causal relationships allows a model to acquire robust knowledge that holds beyond the distribution of a set of training tasks or the observed data~\cite{scholkopf2021toward}. In addition, learning a causal model requires fewer examples to adapt to new environments~\cite{scholkopf2021toward}. \looseness=-1

There has been much research using causal inference to improve OOD generalization. For instance, in social media, \citet{pryzant2018deconfounded} induce a lexicon that is helpful for target label prediction yet uncorrelated to a set of confounding variables, and \citet{Saha2019ASM} perform propensity score-based causal analysis social media posts for evaluating the causal effect of psychiatric medications.
In text classification, \citet{veitch2021counterfactual} use the regularization term Maximum Mean Discrepancy to enforce the prediction of model invariance to the change of non-causal features. \citet{landeiro2016robust} use Pearl’s backdoor adjustment \cite{pearl2000models} to control confounding variables when training a text classifier.

\subsection{Training Techniques}\label{sec:train_tech}
OOD generalization is also addressed using dedicated training objects (strategy). We discuss three NLP-related methods from domain shift, invariance risk, and tuning technique.

\noindent\textbf{Distributionally Robust Optimization (DRO)} aims to learn a model on the worst-case distribution scenario (domain) while expected to generalize well on test data. To improve the worst-case domain, \citet{sagawa2019distributionally} propose a group DRO method that requires explicit group annotation of samples.
Methods based on group DRO and its variants are recently applied in NLP task, such as NLI~\cite{sagawa2019distributionally,liu2021just}, machine translation~\cite{zhou2021distributionally}, spoken language understanding~\cite{broscheitdistributionally}, and toxicity detection~\cite{michel2020modeling}. For example, \citet{oren2019distributionally} design a DRO procedure for generative modeling that minimizes the simulated worst-case distribution scenario over the mixture of topics. Others propose the modified group DROs \cite{sagawa2019distributionally,liu2021just}, and apply in NLI. \citet{zhou2021contrastive} consider the worst-case with language pairs to optimize the multilingual neural machine translation.

% Typically, DRO defines the uncertainty set $\mathcal{Q}$ as s divergence ball around the training distribution over data $(x,y)$. However, $\mathcal{Q}$ is considered too pessimistic when the radius is large. Group DRO method is proposed by giving $\mathcal{Q}$ a wider radius but with fewer degrees of freedom (shift over group instead of $(x,y)$)~\cite{sagawa2019distributionally}, such as label shift or domain shift.

% such as generation task~\cite{oren2019distributionally}, NLI~\cite{sagawa2019distributionally,liu2021just}, MT~\cite{zhou2021distributionally}, spoken language understanding~\cite{broscheitdistributionally}, toxicity detection~\cite{michel2020modeling}.

% \subsubsection{Invariance Risk Minization (IRM)}
\noindent\textbf{Invariance Risk Minimization (IRM)} Different from DRO, which focuses on domain shift robustness, IRM methods focus on learning invariant representations.
IRM \cite{arjovsky2019invariant} is a recently proposed learning paradigm that estimates non-linear, invariant, causal predictors from multiple training environments for improving OOD generalization. It has several advantages. For example, it does not need extra knowledge to manipulate the original data (e.g., human intervention or rule-based method) and extra large computation. Existing work has studied the IRM and its variants in NLP. \citet{choe2020empirical} investigate IRM on synthetic settings and simple MLP and machine larning models in sentiment analysis. \citet{dranker2021irm} study OOD generalization for NLI by IRM, in which environments are constructed by ensuring whether the dataset and bias are synthetic or naturalistic. 
\citet{peyrard2021invariant} propose a language framework based on IRM-games \cite{ahuja2020invariant} for learning invariant representations that generalize better across multiple environments. 
%It performs well in removing structured noise, ignoring spurious correlations, and OOD generalizations, which show the promise in mitigating spurious correlations and biases in models.

% A general predictor $w \circ \Phi $ for environment $\mathscr{E}$ contains a body encoder $\Phi$ and a classifier $w$. The goal of IRM is to learn an invariant representation that yields the same optimal classifier for all environments.

% IRM has two advantages compared to CDA and meta-learning in domain generation, which are also popular methods. Firstly, it does not need extra knowledge to manipulate the original data (e.g. human intervention or rule-based method). Secondly, 

% Experiments based on the above environments prove that learning complex features are still hard in a naturalistic setting. It is still challenging to apply IRM to real-world NLI scenarios for OOD generalization.

\noindent\textbf{Tuning}
Three popular training approaches for preserving the pre-trained features are reviewed: \textit{prompt tuning}, \textit{adapter tuning}, and \textit{linear probing}.

% Pre-trained features are generalizable across domains \cite{kumar2022fine}, because the pre-trained models are trained on large and diverse corpora. Several studies have focused on improving OOD generalization through pre-trained features. We review three popular training approaches for preserving the pre-trained features, \textit{prompt tuning}, \textit{adapter tuning}, and \textit{linear probing}.
\textit{Adapter tuning}~\cite{rebuffi2017learning,houlsby2019parameter} contains a few task-specific trainable parameters and are injected between layers of frozen pre-trained models. Only training the adapter modules can help models achieve competitive performance on various tasks, such as multi-task text classification~\cite{houlsby2019parameter}, NER~\cite{pfeiffer-etal-2020-mad}, multi-tasks QA~\cite{friedman2021single}, and multilingual speech translation~\cite{le-etal-2021-lightweight}.

\textit{Prompt tuning}~\cite{liu2021pre} methods convert the downstream problems into language modeling problems. It adds prompt tokens as the prefix to the questions and converts them as input texts, then use a pre-trained language model to process the input texts for generating the answer sequences. There are two variations of prompt tokens, hard prompt tokens, and soft prompt tokens. Tuning hard prompt tokens requires fine-tuning the pre-trained models \cite{petroni2019language,cui2021template}. Tuning soft prompt tokens only need to fine-tune the prompt tokens, thus preserving the pre-trained features \cite{Li2021PrefixTuningOC,lester2021power,qin-eisner-2021-learning,liu2021p}. Soft prompt tuning is helpful for a wide range of cross-domain tasks, such as NER \cite{chen2021lightner,chen2022few}, text classification \cite{Gao2021MakingPL,Zhong2021AdaptingLM,utama2021avoiding}, table-to-text \cite{Li2021PrefixTuningOC}, QA and paraphrase detection \cite{lester2021power} and NLU \cite{liu2021p}.

\textit{Linear probing}~\cite{liu-gardner-belinkov-peters-smith:2019:NAACL} fine-tunes the top layers while remains the lower layers frozen. Compared to full fine-tuning, linear probing performs better for OOD generalization but reaches lower accuracy on IID data. \citet{kumar2022fine} propose a two-step strategy, which first trains the model with linear probing and then performs fine-tuning (LP-FT). This approach has been theoretically proven to improve both in-domain and OOD performance for deep neural models.

\section{Evaluations}

\noindent\textbf{Unified Evaluations} Although addressing different problems discussed in Sec.\ref{sec:scope}, the methods in Sec.\ref{sec:method} share a large degree of similarities. For example, the invariant feature learning methods have been considered for addressing both machine translation and NLI problems, and the counterfactual (data augmentation) method has been used to solve spurious feature and annotation artifacts issues. Such underlying connections arise from the shared OOD generalization root of such problems, which enables us to design a unified evaluation for PLMs under the same experimental conditions. 

We conclude that existing evaluations \cite{sugawara2018makes,kaushik2018much,min2019compositional,gardner2020evaluating} for OOD generalization of NLP usually keep an eye on only one \cite{srivastava2020robustness,wang2021robustness,howard2022neurocounterfactuals} or a few tasks~\cite{kaushik2019learning,kaushik2020explaining,srivastava2020robustness}, which do not adequately capture the limitations of existing models \cite{tu2020empirical,ribeiro2020beyond}, with the exception of the recent emerging GLUE-X leaderboard \cite{yang2022glue}. However, GLUE-X only includes moderate-sized PLMs, such as BERT-large and T5-large, the evaluation of large-scale language models (e.g., Turing and InstructGPT) is still lacking in the literature, given its crucial positions in research.

\noindent\textbf{Datasets} To provide quick access to existing datasets and facilitate future research, we present label-sharing datasets for each task that can be used for conducting zero-shot cross-domain tests directly in the right-hand column of Tables 1 and 2. Meanwhile, we notice that the unified OOD datasets and the statistic of the distribution shift between them are still lacking exploration. The full list of datasets can be found in Appendix A.
%How Much Reading Does Reading Comprehension Require? A Critical Investigation of Popular Benchmarks
%评测

%What makes reading comprehension questions easier?
%评测

%Beat the AI: Investigating Adversarial Human Annotation for Reading Comprehension
%评测
%training on adversarially collected samples

%Why machine reading comprehension models learn shortcuts?
%评测

%Pay Attention to the Ending: Strong Neural Baselines for the ROC Story Cloze Task
%评测

%Compositional questions do not necessitate multi-hop reasoning.
%评测
%Models in the Loop: Aiding Crowdworkers with Generative Annotation Assistants

%Extending the scope of out-of-domain: Examining qa models in multiple subdomains.
%评测

\section{Future Directions and Conclusion}

We consider multiple promising directions for improving the OOD robustness from three perspectives: (1) enhancing the learning of such salient \textbf{causal features}, either by the help of human guidance \cite{kaushik2019learning,lu2022rationale} via human-in-the-loop, or through psychologically inspired neural structures \cite{chowdhery2022palm}, can be worth consideration; (2) \textbf{data-centric AI:} both the selection of training data and the careful design of prompt learning have shown the effectiveness in domain generalization recently \cite{chen2022knowprompt}. In line with prior work, the distribution alignment method acting as a general regularization can be seen as a promising direction for domain generalization in text classification; (3) \textbf{increasing parameters size of PLMs:} the emerging ability of large-scale language models holds a huge potential to demonstrate relatively high accuracy in the evaluation of domain generalization with the help of reinforcement learning from human feedback.

This paper presents the first attempt at categorising OOD robustness into data- and model-level challenges for reflecting the brittleness of current methods, highlighting the importance of OOD robustness and providing necessary indices as quick access to existing work and materials. This is an ongoing work, and the discussion about LLMs will be added in the next version.

\section{Limitation}
When we categorise OOD-related NLP work, we only focus on the text classification tasks, which can be extended to text generalization tasks, such as machine translation and summarization. Moreover, the literature on domain generalization and domain adaptation has not been distinguished in this work.

\bibliography{anthology}

\begin{thebibliography}{253}
\expandafter\ifx\csname natexlab\endcsname\relax\def\natexlab#1{#1}\fi

\bibitem[{Ahuja et~al.(2020)Ahuja, Shanmugam, Varshney, and
  Dhurandhar}]{ahuja2020invariant}
Kartik Ahuja, Karthikeyan Shanmugam, Kush Varshney, and Amit Dhurandhar. 2020.
\newblock Invariant risk minimization games.
\newblock In \emph{International Conference on Machine Learning}, pages
  145--155. PMLR.

\bibitem[{Andreas(2020)}]{andreas2020good}
Jacob Andreas. 2020.
\newblock Good-enough compositional data augmentation.
\newblock In \emph{Proceedings of the 58th Annual Meeting of the Association
  for Computational Linguistics}, pages 7556--7566.

\bibitem[{Arivazhagan et~al.(2019)Arivazhagan, Bapna, Firat, Aharoni, Johnson,
  and Macherey}]{arivazhagan2019missing}
Naveen Arivazhagan, Ankur Bapna, Orhan Firat, Roee Aharoni, Melvin Johnson, and
  Wolfgang Macherey. 2019.
\newblock The missing ingredient in zero-shot neural machine translation.
\newblock \emph{arXiv preprint arXiv:1903.07091}.

\bibitem[{Arjovsky et~al.(2019)Arjovsky, Bottou, Gulrajani, and
  Lopez-Paz}]{arjovsky2019invariant}
Martin Arjovsky, L{\'e}on Bottou, Ishaan Gulrajani, and David Lopez-Paz. 2019.
\newblock Invariant risk minimization.
\newblock \emph{arXiv preprint arXiv:1907.02893}.

\bibitem[{Arora et~al.(2021)Arora, Huang, and He}]{arora2021types}
Udit Arora, William Huang, and He~He. 2021.
\newblock Types of out-of-distribution texts and how to detect them.
\newblock In \emph{Proceedings of the 2021 Conference on Empirical Methods in
  Natural Language Processing}, pages 10687--10701.

\bibitem[{Bartolo et~al.(2020)Bartolo, Roberts, Welbl, Riedel, and
  Stenetorp}]{bartolo2020beat}
Max Bartolo, Alastair Roberts, Johannes Welbl, Sebastian Riedel, and Pontus
  Stenetorp. 2020.
\newblock Beat the ai: Investigating adversarial human annotation for reading
  comprehension.
\newblock \emph{Transactions of the Association for Computational Linguistics},
  8:662--678.

\bibitem[{Bartolo et~al.(2021{\natexlab{a}})Bartolo, Thrush, Jia, Riedel,
  Stenetorp, and Kiela}]{bartolo2021improving}
Max Bartolo, Tristan Thrush, Robin Jia, Sebastian Riedel, Pontus Stenetorp, and
  Douwe Kiela. 2021{\natexlab{a}}.
\newblock Improving question answering model robustness with synthetic
  adversarial data generation.
\newblock In \emph{Proceedings of the 2021 Conference on Empirical Methods in
  Natural Language Processing}, pages 8830--8848.

\bibitem[{Bartolo et~al.(2021{\natexlab{b}})Bartolo, Thrush, Riedel, Stenetorp,
  Jia, and Kiela}]{bartolo2021models}
Max Bartolo, Tristan Thrush, Sebastian Riedel, Pontus Stenetorp, Robin Jia, and
  Douwe Kiela. 2021{\natexlab{b}}.
\newblock Models in the loop: Aiding crowdworkers with generative annotation
  assistants.
\newblock \emph{arXiv preprint arXiv:2112.09062}.

\bibitem[{Bastings et~al.(2021)Bastings, Ebert, Zablotskaia, Sandholm, and
  Filippova}]{bastings2021will}
Jasmijn Bastings, Sebastian Ebert, Polina Zablotskaia, Anders Sandholm, and
  Katja Filippova. 2021.
\newblock " will you find these shortcuts?" a protocol for evaluating the
  faithfulness of input salience methods for text classification.
\newblock \emph{arXiv preprint arXiv:2111.07367}.

\bibitem[{Bayer et~al.(2021)Bayer, Kaufhold, and Reuter}]{bayer2021survey}
Markus Bayer, Marc-Andr{\'e} Kaufhold, and Christian Reuter. 2021.
\newblock A survey on data augmentation for text classification.
\newblock \emph{arXiv preprint arXiv:2107.03158}.

\bibitem[{Belinkov and Bisk(2017)}]{belinkov2017synthetic}
Yonatan Belinkov and Yonatan Bisk. 2017.
\newblock Synthetic and natural noise both break neural machine translation.
\newblock \emph{arXiv preprint arXiv:1711.02173}.

\bibitem[{Ben-David et~al.(2010)Ben-David, Blitzer, Crammer, Kulesza, Pereira,
  and Vaughan}]{ben2010theory}
Shai Ben-David, John Blitzer, Koby Crammer, Alex Kulesza, Fernando Pereira, and
  Jennifer~Wortman Vaughan. 2010.
\newblock A theory of learning from different domains.
\newblock \emph{Machine learning}, 79(1):151--175.

\bibitem[{Blitzer et~al.(2006)Blitzer, McDonald, and
  Pereira}]{blitzer2006domain}
John Blitzer, Ryan McDonald, and Fernando Pereira. 2006.
\newblock Domain adaptation with structural correspondence learning.
\newblock In \emph{Proceedings of the 2006 conference on empirical methods in
  natural language processing}, pages 120--128.

\bibitem[{Bogin et~al.(2021)Bogin, Subramanian, Gardner, and
  Berant}]{bogin2021latent}
Ben Bogin, Sanjay Subramanian, Matt Gardner, and Jonathan Berant. 2021.
\newblock Latent compositional representations improve systematic
  generalization in grounded question answering.
\newblock \emph{Transactions of the Association for Computational Linguistics},
  9:195--210.

\bibitem[{Bolukbasi et~al.(2016)Bolukbasi, Chang, Zou, Saligrama, and
  Kalai}]{bolukbasi2016man}
Tolga Bolukbasi, Kai-Wei Chang, James~Y Zou, Venkatesh Saligrama, and Adam~T
  Kalai. 2016.
\newblock Man is to computer programmer as woman is to homemaker? debiasing
  word embeddings.
\newblock \emph{Advances in neural information processing systems}, 29.

\bibitem[{Broscheit et~al.()Broscheit, Do, and
  Gaspers}]{broscheitdistributionally}
Samuel Broscheit, Quynh Do, and Judith Gaspers.
\newblock Distributionally robust finetuning bert for covariate drift in spoken
  language understanding.

\bibitem[{Brown et~al.(2020)Brown, Mann, Ryder, Subbiah, Kaplan, Dhariwal,
  Neelakantan, Shyam, Sastry, Askell et~al.}]{brown2020language}
Tom Brown, Benjamin Mann, Nick Ryder, Melanie Subbiah, Jared~D Kaplan, Prafulla
  Dhariwal, Arvind Neelakantan, Pranav Shyam, Girish Sastry, Amanda Askell,
  et~al. 2020.
\newblock Language models are few-shot learners.
\newblock \emph{Advances in neural information processing systems},
  33:1877--1901.

\bibitem[{Cai et~al.(2017)Cai, Tu, and Gimpel}]{cai2017pay}
Zheng Cai, Lifu Tu, and Kevin Gimpel. 2017.
\newblock Pay attention to the ending: Strong neural baselines for the roc
  story cloze task.
\newblock In \emph{Proceedings of the 55th Annual Meeting of the Association
  for Computational Linguistics (Volume 2: Short Papers)}, pages 616--622.

\bibitem[{Caliskan et~al.(2017)Caliskan, Bryson, and
  Narayanan}]{caliskan2017semantics}
Aylin Caliskan, Joanna~J Bryson, and Arvind Narayanan. 2017.
\newblock Semantics derived automatically from language corpora contain
  human-like biases.
\newblock \emph{Science}, 356(6334):183--186.

\bibitem[{Chen et~al.(2022{\natexlab{a}})Chen, He, Narasimhan, and
  Chen}]{chen2022can}
Howard Chen, Jacqueline He, Karthik Narasimhan, and Danqi Chen.
  2022{\natexlab{a}}.
\newblock Can rationalization improve robustness?
\newblock \emph{arXiv preprint arXiv:2204.11790}.

\bibitem[{Chen et~al.(2021{\natexlab{a}})Chen, Tam, Raffel, Bansal, and
  Yang}]{chen2021empirical}
Jiaao Chen, Derek Tam, Colin Raffel, Mohit Bansal, and Diyi Yang.
  2021{\natexlab{a}}.
\newblock An empirical survey of data augmentation for limited data learning in
  nlp.
\newblock \emph{arXiv preprint arXiv:2106.07499}.

\bibitem[{Chen et~al.(2022{\natexlab{b}})Chen, Liu, Lin, Han, and
  Sun}]{chen2022few}
Jiawei Chen, Qing Liu, Hongyu Lin, Xianpei Han, and Le~Sun. 2022{\natexlab{b}}.
\newblock Few-shot named entity recognition with self-describing networks.
\newblock \emph{arXiv preprint arXiv:2203.12252}.

\bibitem[{Chen et~al.(2021{\natexlab{b}})Chen, Aguilar, Neves, and
  Solorio}]{chen2021data}
Shuguang Chen, Gustavo Aguilar, Leonardo Neves, and Thamar Solorio.
  2021{\natexlab{b}}.
\newblock Data augmentation for cross-domain named entity recognition.
\newblock \emph{arXiv preprint arXiv:2109.01758}.

\bibitem[{Chen et~al.(2021{\natexlab{c}})Chen, Zhang, Li, Xie, Deng, Tan,
  Huang, Si, and Chen}]{chen2021lightner}
Xiang Chen, Ningyu Zhang, Lei Li, Xin Xie, Shumin Deng, Chuanqi Tan, Fei Huang,
  Luo Si, and Huajun Chen. 2021{\natexlab{c}}.
\newblock Lightner: A lightweight generative framework with prompt-guided
  attention for low-resource ner.
\newblock \emph{arXiv preprint arXiv:2109.00720}.

\bibitem[{Chen et~al.(2022{\natexlab{c}})Chen, Zhang, Xie, Deng, Yao, Tan,
  Huang, Si, and Chen}]{chen2022knowprompt}
Xiang Chen, Ningyu Zhang, Xin Xie, Shumin Deng, Yunzhi Yao, Chuanqi Tan, Fei
  Huang, Luo Si, and Huajun Chen. 2022{\natexlab{c}}.
\newblock Knowprompt: Knowledge-aware prompt-tuning with synergistic
  optimization for relation extraction.
\newblock In \emph{Proceedings of the ACM Web Conference 2022}, pages
  2778--2788.

\bibitem[{Chen and Cardie(2018)}]{chen2018multinomial}
Xilun Chen and Claire Cardie. 2018.
\newblock Multinomial adversarial networks for multi-domain text
  classification.
\newblock \emph{arXiv preprint arXiv:1802.05694}.

\bibitem[{Chen et~al.(2020)Chen, Liang, Yu, Song, and
  Zhou}]{chen2020compositional}
Xinyun Chen, Chen Liang, Adams~Wei Yu, Dawn Song, and Denny Zhou. 2020.
\newblock Compositional generalization via neural-symbolic stack machines.
\newblock \emph{Advances in Neural Information Processing Systems},
  33:1690--1701.

\bibitem[{Cheng et~al.(2019)Cheng, Jiang, and
  Macherey}]{cheng-etal-2019-robust}
Yong Cheng, Lu~Jiang, and Wolfgang Macherey. 2019.
\newblock \href {https://doi.org/10.18653/v1/P19-1425} {Robust neural machine
  translation with doubly adversarial inputs}.
\newblock In \emph{Proceedings of the 57th Annual Meeting of the Association
  for Computational Linguistics}, pages 4324--4333, Florence, Italy.
  Association for Computational Linguistics.

\bibitem[{Choe et~al.(2020)Choe, Ham, and Park}]{choe2020empirical}
Yo~Joong Choe, Jiyeon Ham, and Kyubyong Park. 2020.
\newblock An empirical study of invariant risk minimization.
\newblock \emph{arXiv preprint arXiv:2004.05007}.

\bibitem[{Choubey et~al.(2021)Choubey, Currey, Mathur, and
  Dinu}]{choubey2021improving}
Prafulla~Kumar Choubey, Anna Currey, Prashant Mathur, and Georgiana Dinu. 2021.
\newblock Improving gender translation accuracy with filtered self-training.
\newblock \emph{arXiv preprint arXiv:2104.07695}.

\bibitem[{Chowdhery et~al.(2022)Chowdhery, Narang, Devlin, Bosma, Mishra,
  Roberts, Barham, Chung, Sutton, Gehrmann et~al.}]{chowdhery2022palm}
Aakanksha Chowdhery, Sharan Narang, Jacob Devlin, Maarten Bosma, Gaurav Mishra,
  Adam Roberts, Paul Barham, Hyung~Won Chung, Charles Sutton, Sebastian
  Gehrmann, et~al. 2022.
\newblock Palm: Scaling language modeling with pathways.
\newblock \emph{arXiv preprint arXiv:2204.02311}.

\bibitem[{Cobbe et~al.(2021)Cobbe, Kosaraju, Bavarian, Hilton, Nakano, Hesse,
  and Schulman}]{cobbe2021training}
Karl Cobbe, Vineet Kosaraju, Mohammad Bavarian, Jacob Hilton, Reiichiro Nakano,
  Christopher Hesse, and John Schulman. 2021.
\newblock Training verifiers to solve math word problems.
\newblock \emph{arXiv preprint arXiv:2110.14168}.

\bibitem[{Cui et~al.(2021)Cui, Wu, Liu, Yang, and Zhang}]{cui2021template}
Leyang Cui, Yu~Wu, Jian Liu, Sen Yang, and Yue Zhang. 2021.
\newblock Template-based named entity recognition using bart.
\newblock \emph{arXiv preprint arXiv:2106.01760}.

\bibitem[{Czarnowska et~al.(2019)Czarnowska, Ruder, Grave, Cotterell, and
  Copestake}]{czarnowska2019don}
Paula Czarnowska, Sebastian Ruder, {\'E}douard Grave, Ryan Cotterell, and Ann
  Copestake. 2019.
\newblock Don’t forget the long tail! a comprehensive analysis of
  morphological generalization in bilingual lexicon induction.
\newblock In \emph{Proceedings of the 2019 Conference on Empirical Methods in
  Natural Language Processing and the 9th International Joint Conference on
  Natural Language Processing (EMNLP-IJCNLP)}, pages 974--983.

\bibitem[{Dankers et~al.(2021)Dankers, Bruni, and Hupkes}]{dankers2021paradox}
Verna Dankers, Elia Bruni, and Dieuwke Hupkes. 2021.
\newblock The paradox of the compositionality of natural language: a neural
  machine translation case study.
\newblock \emph{arXiv preprint arXiv:2108.05885}.

\bibitem[{Das et~al.(2021)Das, Katiyar, Passonneau, and
  Zhang}]{das2021container}
Sarkar Snigdha~Sarathi Das, Arzoo Katiyar, Rebecca~J Passonneau, and Rui Zhang.
  2021.
\newblock Container: Few-shot named entity recognition via contrastive
  learning.
\newblock \emph{arXiv preprint arXiv:2109.07589}.

\bibitem[{Daum{\'e}~III(2009)}]{daume2009frustratingly}
Hal Daum{\'e}~III. 2009.
\newblock Frustratingly easy domain adaptation.
\newblock \emph{arXiv preprint arXiv:0907.1815}.

\bibitem[{Devlin et~al.(2018)Devlin, Chang, Lee, and
  Toutanova}]{devlin2018bert}
Jacob Devlin, Ming-Wei Chang, Kenton Lee, and Kristina Toutanova. 2018.
\newblock Bert: Pre-training of deep bidirectional transformers for language
  understanding.
\newblock \emph{arXiv preprint arXiv:1810.04805}.

\bibitem[{Diaz et~al.(2018)Diaz, Johnson, Lazar, Piper, and Gergle}]{agebias}
Mark Diaz, Isaac Johnson, Amanda Lazar, Anne~Marie Piper, and Darren Gergle.
  2018.
\newblock Addressing age-related bias in sentiment analysis.
\newblock In \emph{Proceedings of the 2018 CHI Conference on Human Factors in
  Computing Systems}. Association for Computing Machinery.

\bibitem[{Dixon et~al.(2018)Dixon, Li, Sorensen, Thain, and
  Vasserman}]{Dixon2018MeasuringAM}
Lucas Dixon, John Li, Jeffrey~Scott Sorensen, Nithum Thain, and Lucy Vasserman.
  2018.
\newblock Measuring and mitigating unintended bias in text classification.
\newblock \emph{Proceedings of the 2018 AAAI/ACM Conference on AI, Ethics, and
  Society}.

\bibitem[{Dong and Lapata(2016)}]{dong2016language}
Li~Dong and Mirella Lapata. 2016.
\newblock Language to logical form with neural attention.
\newblock In \emph{Proceedings of the 54th Annual Meeting of the Association
  for Computational Linguistics (Volume 1: Long Papers)}, pages 33--43.

\bibitem[{Dong and Lapata(2018)}]{dong2018coarse}
Li~Dong and Mirella Lapata. 2018.
\newblock Coarse-to-fine decoding for neural semantic parsing.
\newblock In \emph{Proceedings of the 56th Annual Meeting of the Association
  for Computational Linguistics (ACL 2018)}, pages 731--742.

\bibitem[{Dranker et~al.(2021)Dranker, He, and Belinkov}]{dranker2021irm}
Yana Dranker, He~He, and Yonatan Belinkov. 2021.
\newblock Irm---when it works and when it doesn't: A test case of natural
  language inference.
\newblock \emph{Advances in Neural Information Processing Systems}, 34.

\bibitem[{Drori et~al.(2021)Drori, Tran, Wang, Cheng, Liu, Tang, Ke, Singh,
  Patti, Lynch et~al.}]{drori2021neural}
Iddo Drori, Sunny Tran, Roman Wang, Newman Cheng, Kevin Liu, Leonard Tang,
  Elizabeth Ke, Nikhil Singh, Taylor~L Patti, Jayson Lynch, et~al. 2021.
\newblock A neural network solves and generates mathematics problems by program
  synthesis: Calculus, differential equations, linear algebra, and more.
\newblock \emph{arXiv preprint arXiv:2112.15594}.

\bibitem[{Du et~al.(2020)Du, Sun, Wang, Qi, and Liao}]{du2020adversarial}
Chunning Du, Haifeng Sun, Jingyu Wang, Qi~Qi, and Jianxin Liao. 2020.
\newblock Adversarial and domain-aware bert for cross-domain sentiment
  analysis.
\newblock In \emph{Proceedings of the 58th annual meeting of the Association
  for Computational Linguistics}, pages 4019--4028.

\bibitem[{Du et~al.(2021{\natexlab{a}})Du, Manjunatha, Jain, Deshpande,
  Dernoncourt, Gu, Sun, and Hu}]{du2021towards}
Mengnan Du, Varun Manjunatha, Rajiv Jain, Ruchi Deshpande, Franck Dernoncourt,
  Jiuxiang Gu, Tong Sun, and Xia Hu. 2021{\natexlab{a}}.
\newblock Towards interpreting and mitigating shortcut learning behavior of nlu
  models.
\newblock In \emph{Proceedings of the 2021 Conference of the North American
  Chapter of the Association for Computational Linguistics: Human Language
  Technologies}, pages 915--929.

\bibitem[{Du et~al.(2021{\natexlab{b}})Du, Wang, Feng, Pan, Qin, Xu, and
  Wang}]{du2021adarnn}
Yuntao Du, Jindong Wang, Wenjie Feng, Sinno Pan, Tao Qin, Renjun Xu, and
  Chongjun Wang. 2021{\natexlab{b}}.
\newblock Adarnn: Adaptive learning and forecasting of time series.
\newblock In \emph{Proceedings of the 30th ACM International Conference on
  Information \& Knowledge Management}, pages 402--411.

\bibitem[{Eriguchi et~al.(2018)Eriguchi, Johnson, Firat, Kazawa, and
  Macherey}]{eriguchi2018zero}
Akiko Eriguchi, Melvin Johnson, Orhan Firat, Hideto Kazawa, and Wolfgang
  Macherey. 2018.
\newblock Zero-shot cross-lingual classification using multilingual neural
  machine translation.
\newblock \emph{arXiv preprint arXiv:1809.04686}.

\bibitem[{Feng et~al.(2019)Feng, Wallace, and Boyd-Graber}]{feng2019misleading}
Shi Feng, Eric Wallace, and Jordan Boyd-Graber. 2019.
\newblock Misleading failures of partial-input baselines.
\newblock In \emph{Proceedings of the 57th Annual Meeting of the Association
  for Computational Linguistics}, pages 5533--5538.

\bibitem[{Feng et~al.(2021)Feng, Gangal, Wei, Chandar, Vosoughi, Mitamura, and
  Hovy}]{feng2021survey}
Steven~Y Feng, Varun Gangal, Jason Wei, Sarath Chandar, Soroush Vosoughi,
  Teruko Mitamura, and Eduard Hovy. 2021.
\newblock A survey of data augmentation approaches for nlp.
\newblock In \emph{Findings of the Association for Computational Linguistics:
  ACL-IJCNLP 2021}, pages 968--988.

\bibitem[{Fort et~al.(2021)Fort, Ren, and Lakshminarayanan}]{fort2021exploring}
Stanislav Fort, Jie Ren, and Balaji Lakshminarayanan. 2021.
\newblock Exploring the limits of out-of-distribution detection.
\newblock \emph{Advances in Neural Information Processing Systems}, 34.

\bibitem[{Fried et~al.(2019)Fried, Kitaev, and Klein}]{fried-etal-2019-cross}
Daniel Fried, Nikita Kitaev, and Dan Klein. 2019.
\newblock \href {https://doi.org/10.18653/v1/P19-1031} {Cross-domain
  generalization of neural constituency parsers}.
\newblock In \emph{Proceedings of the 57th Annual Meeting of the Association
  for Computational Linguistics}, pages 323--330, Florence, Italy. Association
  for Computational Linguistics.

\bibitem[{Friedman et~al.(2021)Friedman, Dodge, and Chen}]{friedman2021single}
Dan Friedman, Ben Dodge, and Danqi Chen. 2021.
\newblock Single-dataset experts for multi-dataset question answering.
\newblock In \emph{Proceedings of the 2021 Conference on Empirical Methods in
  Natural Language Processing}, pages 6128--6137.

\bibitem[{Furrer et~al.(2020)Furrer, van Zee, Scales, and
  Sch{\"a}rli}]{furrer2020compositional}
Daniel Furrer, Marc van Zee, Nathan Scales, and Nathanael Sch{\"a}rli. 2020.
\newblock Compositional generalization in semantic parsing: Pre-training vs.
  specialized architectures.
\newblock \emph{arXiv preprint arXiv:2007.08970}.

\bibitem[{Gagnon-Audet et~al.(2022)Gagnon-Audet, Ahuja, Darvishi-Bayazi, Dumas,
  and Rish}]{gagnon2022woods}
Jean-Christophe Gagnon-Audet, Kartik Ahuja, Mohammad-Javad Darvishi-Bayazi,
  Guillaume Dumas, and Irina Rish. 2022.
\newblock Woods: Benchmarks for out-of-distribution generalization in time
  series tasks.
\newblock \emph{arXiv preprint arXiv:2203.09978}.

\bibitem[{Ganin et~al.(2016)Ganin, Ustinova, Ajakan, Germain, Larochelle,
  Laviolette, Marchand, and Lempitsky}]{ganin2016domain}
Yaroslav Ganin, Evgeniya Ustinova, Hana Ajakan, Pascal Germain, Hugo
  Larochelle, Fran{\c{c}}ois Laviolette, Mario Marchand, and Victor Lempitsky.
  2016.
\newblock Domain-adversarial training of neural networks.
\newblock \emph{The journal of machine learning research}, 17(1):2096--2030.

\bibitem[{Gao et~al.(2021)Gao, Fisch, and Chen}]{Gao2021MakingPL}
Tianyu Gao, Adam Fisch, and Danqi Chen. 2021.
\newblock Making pre-trained language models better few-shot learners.
\newblock \emph{ArXiv}, abs/2012.15723.

\bibitem[{Gardner et~al.(2020)Gardner, Artzi, Basmov, Berant, Bogin, Chen,
  Dasigi, Dua, Elazar, Gottumukkala et~al.}]{gardner2020evaluating}
Matt Gardner, Yoav Artzi, Victoria Basmov, Jonathan Berant, Ben Bogin, Sihao
  Chen, Pradeep Dasigi, Dheeru Dua, Yanai Elazar, Ananth Gottumukkala, et~al.
  2020.
\newblock Evaluating models’ local decision boundaries via contrast sets.
\newblock In \emph{Findings of the Association for Computational Linguistics:
  EMNLP 2020}, pages 1307--1323.

\bibitem[{Gardner et~al.(2021)Gardner, Merrill, Dodge, Peters, Ross, Singh, and
  Smith}]{gardner2021competency}
Matt Gardner, William Merrill, Jesse Dodge, Matthew~E Peters, Alexis Ross,
  Sameer Singh, and Noah~A Smith. 2021.
\newblock Competency problems: On finding and removing artifacts in language
  data.
\newblock In \emph{Proceedings of the 2021 Conference on Empirical Methods in
  Natural Language Processing}, pages 1801--1813.

\bibitem[{Geirhos et~al.(2020)Geirhos, Jacobsen, Michaelis, Zemel, Brendel,
  Bethge, and Wichmann}]{geirhos2020shortcut}
Robert Geirhos, J{\"o}rn-Henrik Jacobsen, Claudio Michaelis, Richard Zemel,
  Wieland Brendel, Matthias Bethge, and Felix~A Wichmann. 2020.
\newblock Shortcut learning in deep neural networks.
\newblock \emph{Nature Machine Intelligence}, 2(11):665--673.

\bibitem[{Ghaddar and Langlais(2017)}]{ghaddar2017winer}
Abbas Ghaddar and Philippe Langlais. 2017.
\newblock Winer: A wikipedia annotated corpus for named entity recognition.
\newblock In \emph{Proceedings of the Eighth International Joint Conference on
  Natural Language Processing (Volume 1: Long Papers)}, pages 413--422.

\bibitem[{Gokhale et~al.(2022)Gokhale, Mishra, Luo, Sachdeva, and
  Baral}]{gokhale2022generalized}
Tejas Gokhale, Swaroop Mishra, Man Luo, Bhavdeep~Singh Sachdeva, and Chitta
  Baral. 2022.
\newblock Generalized but not robust? comparing the effects of data
  modification methods on out-of-domain generalization and adversarial
  robustness.
\newblock \emph{arXiv preprint arXiv:2203.07653}.

\bibitem[{Gonen and Goldberg(2019)}]{gonen2019lipstick}
Hila Gonen and Yoav Goldberg. 2019.
\newblock Lipstick on a pig: Debiasing methods cover up systematic gender
  biases in word embeddings but do not remove them.
\newblock In \emph{Proceedings of the 2019 Conference of the North American
  Chapter of the Association for Computational Linguistics: Human Language
  Technologies, Volume 1 (Long and Short Papers)}, pages 609--614.

\bibitem[{Goodfellow et~al.(2015)Goodfellow, Shlens, and
  Szegedy}]{Goodfellow2015ExplainingAH}
Ian~J. Goodfellow, Jonathon Shlens, and Christian Szegedy. 2015.
\newblock Explaining and harnessing adversarial examples.
\newblock \emph{CoRR}, abs/1412.6572.

\bibitem[{Gordon et~al.(2019)Gordon, Lopez-Paz, Baroni, and
  Bouchacourt}]{gordon2019permutation}
Jonathan Gordon, David Lopez-Paz, Marco Baroni, and Diane Bouchacourt. 2019.
\newblock Permutation equivariant models for compositional generalization in
  language.
\newblock In \emph{International Conference on Learning Representations}.

\bibitem[{Gozalo-Brizuela and Garrido-Merchan(2023)}]{gozalo2023chatgpt}
Roberto Gozalo-Brizuela and Eduardo~C Garrido-Merchan. 2023.
\newblock Chatgpt is not all you need. a state of the art review of large
  generative ai models.
\newblock \emph{arXiv preprint arXiv:2301.04655}.

\bibitem[{Gu et~al.(2021)Gu, Kase, Vanni, Sadler, Liang, Yan, and
  Su}]{gu2021beyond}
Yu~Gu, Sue Kase, Michelle Vanni, Brian Sadler, Percy Liang, Xifeng Yan, and
  Yu~Su. 2021.
\newblock Beyond iid: three levels of generalization for question answering on
  knowledge bases.
\newblock In \emph{Proceedings of the Web Conference 2021}, pages 3477--3488.

\bibitem[{Guo et~al.(2020)Guo, Pasunuru, and Bansal}]{guo2020multi}
Han Guo, Ramakanth Pasunuru, and Mohit Bansal. 2020.
\newblock Multi-source domain adaptation for text classification via
  distancenet-bandits.
\newblock In \emph{Proceedings of the AAAI Conference on Artificial
  Intelligence}, volume~34, pages 7830--7838.

\bibitem[{Gupta et~al.(2022)Gupta, Singh, and Gardner}]{gupta2022structurally}
Shivanshu Gupta, Sameer Singh, and Matt Gardner. 2022.
\newblock Structurally diverse sampling reduces spurious correlations in
  semantic parsing datasets.
\newblock \emph{arXiv preprint arXiv:2203.08445}.

\bibitem[{Gururangan et~al.(2018)Gururangan, Swayamdipta, Levy, Schwartz,
  Bowman, and Smith}]{gururangan2018annotation}
Suchin Gururangan, Swabha Swayamdipta, Omer Levy, Roy Schwartz, Samuel Bowman,
  and Noah~A Smith. 2018.
\newblock Annotation artifacts in natural language inference data.
\newblock In \emph{Proceedings of the 2018 Conference of the North American
  Chapter of the Association for Computational Linguistics: Human Language
  Technologies, Volume 2 (Short Papers)}, pages 107--112.

\bibitem[{Han and Eisenstein(2019)}]{han2019unsupervised}
Xiaochuang Han and Jacob Eisenstein. 2019.
\newblock Unsupervised domain adaptation of contextualized embeddings for
  sequence labeling.
\newblock In \emph{Proceedings of the 2019 Conference on Empirical Methods in
  Natural Language Processing and the 9th International Joint Conference on
  Natural Language Processing (EMNLP-IJCNLP)}, pages 4238--4248.

\bibitem[{Hedderich et~al.(2021)Hedderich, Lange, Adel, Str{\"o}tgen, and
  Klakow}]{hedderich2021survey}
Michael~A Hedderich, Lukas Lange, Heike Adel, Jannik Str{\"o}tgen, and Dietrich
  Klakow. 2021.
\newblock A survey on recent approaches for natural language processing in
  low-resource scenarios.
\newblock In \emph{Proceedings of the 2021 Conference of the North American
  Chapter of the Association for Computational Linguistics: Human Language
  Technologies}, pages 2545--2568.

\bibitem[{Hendrycks et~al.(2021)Hendrycks, Burns, Kadavath, Arora, Basart,
  Tang, Song, and Steinhardt}]{hendrycks2021measuring}
Dan Hendrycks, Collin Burns, Saurav Kadavath, Akul Arora, Steven Basart, Eric
  Tang, Dawn Song, and Jacob Steinhardt. 2021.
\newblock Measuring mathematical problem solving with the math dataset.
\newblock \emph{arXiv preprint arXiv:2103.03874}.

\bibitem[{Hendrycks and Gimpel(2017)}]{hendrycks2017baseline}
Dan Hendrycks and Kevin Gimpel. 2017.
\newblock A baseline for detecting misclassified and out-of-distribution
  examples in neural networks.
\newblock In \emph{International Conference on Learning Representations}.

\bibitem[{Hendrycks et~al.(2020)Hendrycks, Liu, Wallace, Dziedzic, Krishnan,
  and Song}]{hendrycks2020pretrained}
Dan Hendrycks, Xiaoyuan Liu, Eric Wallace, Adam Dziedzic, Rishabh Krishnan, and
  Dawn Song. 2020.
\newblock Pretrained transformers improve out-of-distribution robustness.
\newblock In \emph{Proceedings of the 58th Annual Meeting of the Association
  for Computational Linguistics (ACL 2020)}, pages 2744--2751.

\bibitem[{Hewitt et~al.(2021)Hewitt, Li, Xie, Newman, and
  Liang}]{hewitt2021ensembles}
John Hewitt, Xiang~Lisa Li, Sang~Michael Xie, Benjamin Newman, and Percy Liang.
  2021.
\newblock Ensembles and cocktails: Robust finetuning for natural language
  generation.
\newblock In \emph{NeurIPS 2021 Workshop on Distribution Shifts: Connecting
  Methods and Applications}.

\bibitem[{Holland(1986)}]{Holland1986StatisticsAC}
Paul Holland. 1986.
\newblock Statistics and causal inference.
\newblock \emph{Journal of the American Statistical Association}, 81:945--960.

\bibitem[{Houlsby et~al.(2019)Houlsby, Giurgiu, Jastrzebski, Morrone,
  De~Laroussilhe, Gesmundo, Attariyan, and Gelly}]{houlsby2019parameter}
Neil Houlsby, Andrei Giurgiu, Stanislaw Jastrzebski, Bruna Morrone, Quentin
  De~Laroussilhe, Andrea Gesmundo, Mona Attariyan, and Sylvain Gelly. 2019.
\newblock Parameter-efficient transfer learning for nlp.
\newblock In \emph{International Conference on Machine Learning}, pages
  2790--2799. PMLR.

\bibitem[{Howard et~al.(2022)Howard, Singer, Lal, Choi, and
  Swayamdipta}]{howard2022neurocounterfactuals}
Phillip Howard, Gadi Singer, Vasudev Lal, Yejin Choi, and Swabha Swayamdipta.
  2022.
\newblock Neurocounterfactuals: Beyond minimal-edit counterfactuals for richer
  data augmentation.
\newblock \emph{arXiv preprint arXiv:2210.12365}.

\bibitem[{Hu et~al.(2020)Hu, Ruder, Siddhant, Neubig, Firat, and
  Johnson}]{hu2020xtreme}
Junjie Hu, Sebastian Ruder, Aditya Siddhant, Graham Neubig, Orhan Firat, and
  Melvin Johnson. 2020.
\newblock Xtreme: A massively multilingual multi-task benchmark for evaluating
  cross-lingual generalisation.
\newblock In \emph{International Conference on Machine Learning}, pages
  4411--4421. PMLR.

\bibitem[{Huang et~al.(2021{\natexlab{a}})Huang, Chen, Liu, Xie, Sun, and
  Zhao}]{huang2021nsrl}
Xiusheng Huang, Yubo Chen, Kang Liu, Yuantao Xie, Weijian Sun, and Jun Zhao.
  2021{\natexlab{a}}.
\newblock Nsrl: Named entity recognition with noisy labels via selective review
  learning.
\newblock In \emph{China Conference on Knowledge Graph and Semantic Computing},
  pages 157--170. Springer.

\bibitem[{Huang et~al.(2021{\natexlab{b}})Huang, Fang, Cao, Wang, and
  Liang}]{huang2021dagn}
Yinya Huang, Meng Fang, Yu~Cao, Liwei Wang, and Xiaodan Liang.
  2021{\natexlab{b}}.
\newblock Dagn: Discourse-aware graph network for logical reasoning.
\newblock In \emph{Proceedings of the 2021 Conference of the North American
  Chapter of the Association for Computational Linguistics: Human Language
  Technologies}, pages 5848--5855.

\bibitem[{Hupkes et~al.(2020)Hupkes, Dankers, Mul, and
  Bruni}]{hupkes2020compositionality}
Dieuwke Hupkes, Verna Dankers, Mathijs Mul, and Elia Bruni. 2020.
\newblock Compositionality decomposed: How do neural networks generalise?
\newblock \emph{Journal of Artificial Intelligence Research}, 67:757--795.

\bibitem[{Hupkes et~al.(2022)Hupkes, Giulianelli, Dankers, Artetxe, Elazar,
  Pimentel, Christodoulopoulos, Lasri, Saphra, Sinclair
  et~al.}]{hupkes2022state}
Dieuwke Hupkes, Mario Giulianelli, Verna Dankers, Mikel Artetxe, Yanai Elazar,
  Tiago Pimentel, Christos Christodoulopoulos, Karim Lasri, Naomi Saphra,
  Arabella Sinclair, et~al. 2022.
\newblock State-of-the-art generalisation research in nlp: a taxonomy and
  review.
\newblock \emph{arXiv preprint arXiv:2210.03050}.

\bibitem[{Hutchinson et~al.(2020)Hutchinson, Prabhakaran, Denton, Webster,
  Zhong, and Denuyl}]{hutchinson-etal-2020-social}
Ben Hutchinson, Vinodkumar Prabhakaran, Emily Denton, Kellie Webster, Yu~Zhong,
  and Stephen Denuyl. 2020.
\newblock Social biases in {NLP} models as barriers for persons with
  disabilities.
\newblock In \emph{Proceedings of the 58th Annual Meeting of the Association
  for Computational Linguistics}, Online. Association for Computational
  Linguistics.

\bibitem[{Imbens and Rubin(2015)}]{Imbens2015CausalIF}
Guido Imbens and Donald~B. Rubin. 2015.
\newblock Causal inference for statistics, social, and biomedical sciences: An
  introduction.

\bibitem[{Iyer et~al.(2017)Iyer, Konstas, Cheung, Krishnamurthy, and
  Zettlemoyer}]{iyer2017learning}
Srinivasan Iyer, Ioannis Konstas, Alvin Cheung, Jayant Krishnamurthy, and Luke
  Zettlemoyer. 2017.
\newblock Learning a neural semantic parser from user feedback.
\newblock In \emph{Proceedings of the 55th Annual Meeting of the Association
  for Computational Linguistics (Volume 1: Long Papers)}, pages 963--973.

\bibitem[{Ji et~al.(2020)Ji, Zhang, Duan, Zhang, Chen, and Luo}]{ji2020cross}
Baijun Ji, Zhirui Zhang, Xiangyu Duan, Min Zhang, Boxing Chen, and Weihua Luo.
  2020.
\newblock Cross-lingual pre-training based transfer for zero-shot neural
  machine translation.
\newblock In \emph{Proceedings of the AAAI Conference on Artificial
  Intelligence}, volume~34, pages 115--122.

\bibitem[{Jia et~al.(2019)Jia, Liang, and Zhang}]{jia2019cross}
Chen Jia, Xiaobo Liang, and Yue Zhang. 2019.
\newblock Cross-domain ner using cross-domain language modeling.
\newblock In \emph{Proceedings of the 57th Annual Meeting of the Association
  for Computational Linguistics}, pages 2464--2474.

\bibitem[{Jia and Zhang(2020)}]{jia2020multi}
Chen Jia and Yue Zhang. 2020.
\newblock Multi-cell compositional lstm for ner domain adaptation.
\newblock In \emph{Proceedings of the 58th Annual Meeting of the Association
  for Computational Linguistics}, pages 5906--5917.

\bibitem[{Jiang et~al.(2020)Jiang, He, Chen, Liu, Gao, and
  Zhao}]{jiang2020smart}
Haoming Jiang, Pengcheng He, Weizhu Chen, Xiaodong Liu, Jianfeng Gao, and Tuo
  Zhao. 2020.
\newblock Smart: Robust and efficient fine-tuning for pre-trained natural
  language models through principled regularized optimization.
\newblock In \emph{Proceedings of the 58th Annual Meeting of the Association
  for Computational Linguistics}, pages 2177--2190.

\bibitem[{Johnson et~al.(2017)Johnson, Schuster, Le, Krikun, Wu, Chen, Thorat,
  Vi{\'e}gas, Wattenberg, Corrado, Hughes, and Dean}]{Johnson2017GooglesMN}
Melvin Johnson, Mike Schuster, Quoc~V. Le, Maxim Krikun, Yonghui Wu, Z.~Chen,
  Nikhil Thorat, Fernanda~B. Vi{\'e}gas, Martin Wattenberg, Gregory~S. Corrado,
  Macduff Hughes, and Jeffrey Dean. 2017.
\newblock Google’s multilingual neural machine translation system: Enabling
  zero-shot translation.
\newblock \emph{Transactions of the Association for Computational Linguistics},
  5:339--351.

\bibitem[{Johnson and Zhang(2005)}]{johnson2005high}
Rie Johnson and Tong Zhang. 2005.
\newblock A high-performance semi-supervised learning method for text chunking.
\newblock In \emph{Proceedings of the 43rd Annual Meeting of the Association
  for Computational Linguistics (ACL’05)}, pages 1--9.

\bibitem[{Jung(2022)}]{jung2022machine}
Alexander Jung. 2022.
\newblock \emph{Machine learning: The basics}.
\newblock Springer Nature.

\bibitem[{Kaushik et~al.(2019)Kaushik, Hovy, and Lipton}]{kaushik2019learning}
Divyansh Kaushik, Eduard Hovy, and Zachary Lipton. 2019.
\newblock Learning the difference that makes a difference with
  counterfactually-augmented data.
\newblock In \emph{International Conference on Learning Representations}.

\bibitem[{Kaushik and Lipton(2018)}]{kaushik2018much}
Divyansh Kaushik and Zachary~C Lipton. 2018.
\newblock How much reading does reading comprehension require? a critical
  investigation of popular benchmarks.
\newblock In \emph{Proceedings of the 2018 Conference on Empirical Methods in
  Natural Language Processing}, pages 5010--5015.

\bibitem[{Kaushik et~al.(2020)Kaushik, Setlur, Hovy, and
  Lipton}]{kaushik2020explaining}
Divyansh Kaushik, Amrith Setlur, Eduard~H Hovy, and Zachary~Chase Lipton. 2020.
\newblock Explaining the efficacy of counterfactually augmented data.
\newblock In \emph{International Conference on Learning Representations}.

\bibitem[{Kavumba et~al.(2019)Kavumba, Inoue, Heinzerling, Singh, Reisert, and
  Inui}]{kavumba2019choosing}
Pride Kavumba, Naoya Inoue, Benjamin Heinzerling, Keshav Singh, Paul Reisert,
  and Kentaro Inui. 2019.
\newblock When choosing plausible alternatives, clever hans can be clever.
\newblock In \emph{Proceedings of the First Workshop on Commonsense Inference
  in Natural Language Processing}, pages 33--42.

\bibitem[{Keith et~al.(2020)Keith, Jensen, and O’Connor}]{keith2020text}
Katherine Keith, David Jensen, and Brendan O’Connor. 2020.
\newblock Text and causal inference: A review of using text to remove
  confounding from causal estimates.
\newblock In \emph{Proceedings of the 58th Annual Meeting of the Association
  for Computational Linguistics}, pages 5332--5344.

\bibitem[{Keung et~al.(2019)Keung, Lu, and
  Bhardwaj}]{keung-etal-2019-adversarial}
Phillip Keung, Yichao Lu, and Vikas Bhardwaj. 2019.
\newblock Adversarial learning with contextual embeddings for zero-resource
  cross-lingual classification and {NER}.
\newblock In \emph{Proceedings of the 2019 Conference on Empirical Methods in
  Natural Language Processing and the 9th International Joint Conference on
  Natural Language Processing (EMNLP-IJCNLP)}. Association for Computational
  Linguistics.

\bibitem[{Keysers et~al.(2020)Keysers, Sch{\"a}rli, Scales, Buisman, Furrer,
  Kashubin, Momchev, Sinopalnikov, Stafiniak, Tihon
  et~al.}]{keysers2020measuring}
Daniel Keysers, Nathanael Sch{\"a}rli, Nathan Scales, Hylke Buisman, Daniel
  Furrer, Sergii Kashubin, Nikola Momchev, Danila Sinopalnikov, Lukasz
  Stafiniak, Tibor Tihon, et~al. 2020.
\newblock Measuring compositional generalization: A comprehensive method on
  realistic data.
\newblock In \emph{International Conference on Learning Representations}.

\bibitem[{Khayrallah and Koehn(2018)}]{khayrallah-koehn-2018-impact}
Huda Khayrallah and Philipp Koehn. 2018.
\newblock \href {https://doi.org/10.18653/v1/W18-2709} {On the impact of
  various types of noise on neural machine translation}.
\newblock In \emph{Proceedings of the 2nd Workshop on Neural Machine
  Translation and Generation}, pages 74--83, Melbourne, Australia. Association
  for Computational Linguistics.

\bibitem[{Kim et~al.(2021)Kim, Ravikumar, Ainslie, and
  Onta{\~n}{\'o}n}]{kim2021improving}
Juyong Kim, Pradeep Ravikumar, Joshua Ainslie, and Santiago Onta{\~n}{\'o}n.
  2021.
\newblock Improving compositional generalization in classification tasks via
  structure annotations.
\newblock \emph{arXiv preprint arXiv:2106.10434}.

\bibitem[{Kim and Linzen(2020)}]{kim2020cogs}
Najoung Kim and Tal Linzen. 2020.
\newblock Cogs: A compositional generalization challenge based on semantic
  interpretation.
\newblock In \emph{Proceedings of the 2020 Conference on Empirical Methods in
  Natural Language Processing (EMNLP)}, pages 9087--9105.

\bibitem[{Kim(2021)}]{kim2021sequence}
Yoon Kim. 2021.
\newblock Sequence-to-sequence learning with latent neural grammars.
\newblock \emph{Advances in Neural Information Processing Systems}, 34.

\bibitem[{Kiritchenko and Mohammad(2018)}]{kiritchenko2018examining}
Svetlana Kiritchenko and Saif~M Mohammad. 2018.
\newblock Examining gender and race bias in two hundred sentiment analysis
  systems.
\newblock \emph{arXiv preprint arXiv:1805.04508}.

\bibitem[{Koh et~al.(2021)Koh, Sagawa, Marklund, Xie, Zhang, Balsubramani, Hu,
  Yasunaga, Phillips, Gao et~al.}]{koh2021wilds}
Pang~Wei Koh, Shiori Sagawa, Henrik Marklund, Sang~Michael Xie, Marvin Zhang,
  Akshay Balsubramani, Weihua Hu, Michihiro Yasunaga, Richard~Lanas Phillips,
  Irena Gao, et~al. 2021.
\newblock Wilds: A benchmark of in-the-wild distribution shifts.
\newblock In \emph{International Conference on Machine Learning}, pages
  5637--5664. PMLR.

\bibitem[{Kumar et~al.(2022)Kumar, Raghunathan, Jones, Ma, and
  Liang}]{kumar2022fine}
Ananya Kumar, Aditi Raghunathan, Robbie Jones, Tengyu Ma, and Percy Liang.
  2022.
\newblock Fine-tuning can distort pretrained features and underperform
  out-of-distribution.
\newblock In \emph{International Conference on Learning Representations}.

\bibitem[{Kumar et~al.(2020)Kumar, Choudhary, and Cho}]{kumar2020data}
Varun Kumar, Ashutosh Choudhary, and Eunah Cho. 2020.
\newblock Data augmentation using pre-trained transformer models.
\newblock In \emph{Proceedings of the 2nd Workshop on Life-long Learning for
  Spoken Language Systems}, pages 18--26.

\bibitem[{Lai et~al.(2021)Lai, Zhang, Feng, Huang, and Zhao}]{lai2021machine}
Yuxuan Lai, Chen Zhang, Yansong Feng, Quzhe Huang, and Dongyan Zhao. 2021.
\newblock Why machine reading comprehension models learn shortcuts?
\newblock In \emph{Findings of the Association for Computational Linguistics:
  ACL-IJCNLP 2021}, pages 989--1002.

\bibitem[{Lake and Baroni(2018)}]{lake2018generalization}
Brenden Lake and Marco Baroni. 2018.
\newblock Generalization without systematicity: On the compositional skills of
  sequence-to-sequence recurrent networks.
\newblock In \emph{International conference on machine learning}, pages
  2873--2882. PMLR.

\bibitem[{Lake(2019)}]{lake2019compositional}
Brenden~M Lake. 2019.
\newblock Compositional generalization through meta sequence-to-sequence
  learning.
\newblock \emph{Advances in neural information processing systems}, 32.

\bibitem[{Landeiro and Culotta(2016)}]{landeiro2016robust}
Virgile Landeiro and Aron Culotta. 2016.
\newblock Robust text classification in the presence of confounding bias.
\newblock In \emph{Thirtieth AAAI Conference on Artificial Intelligence}.

\bibitem[{Laparra et~al.(2020)Laparra, Bethard, and
  Miller}]{laparra2020rethinking}
Egoitz Laparra, Steven Bethard, and Timothy~A Miller. 2020.
\newblock Rethinking domain adaptation for machine learning over clinical
  language.
\newblock \emph{JAMIA open}, 3(2):146--150.

\bibitem[{Lazaridou et~al.(2021)Lazaridou, Kuncoro, Gribovskaya, Agrawal,
  Liska, Terzi, Gimenez, de~Masson~d'Autume, Kocisky, Ruder
  et~al.}]{lazaridou2021mind}
Angeliki Lazaridou, Adhi Kuncoro, Elena Gribovskaya, Devang Agrawal, Adam
  Liska, Tayfun Terzi, Mai Gimenez, Cyprien de~Masson~d'Autume, Tomas Kocisky,
  Sebastian Ruder, et~al. 2021.
\newblock Mind the gap: Assessing temporal generalization in neural language
  models.
\newblock \emph{Advances in Neural Information Processing Systems}, 34.

\bibitem[{Le et~al.(2021)Le, Pino, Wang, Gu, Schwab, and
  Besacier}]{le-etal-2021-lightweight}
Hang Le, Juan Pino, Changhan Wang, Jiatao Gu, Didier Schwab, and Laurent
  Besacier. 2021.
\newblock \href {https://doi.org/10.18653/v1/2021.acl-short.103} {Lightweight
  adapter tuning for multilingual speech translation}.
\newblock In \emph{Proceedings of the 59th Annual Meeting of the Association
  for Computational Linguistics and the 11th International Joint Conference on
  Natural Language Processing (Volume 2: Short Papers)}, pages 817--824,
  Online. Association for Computational Linguistics.

\bibitem[{Le~Bras et~al.(2020)Le~Bras, Swayamdipta, Bhagavatula, Zellers,
  Peters, Sabharwal, and Choi}]{le2020adversarial}
Ronan Le~Bras, Swabha Swayamdipta, Chandra Bhagavatula, Rowan Zellers, Matthew
  Peters, Ashish Sabharwal, and Yejin Choi. 2020.
\newblock Adversarial filters of dataset biases.
\newblock In \emph{International Conference on Machine Learning}, pages
  1078--1088. PMLR.

\bibitem[{Lee et~al.(2021)Lee, Agarwal, Kadakia, Pujara, and Ren}]{lee2021good}
Dong-Ho Lee, Mahak Agarwal, Akshen Kadakia, Jay Pujara, and Xiang Ren. 2021.
\newblock Good examples make a faster learner: Simple demonstration-based
  learning for low-resource ner.
\newblock \emph{arXiv preprint arXiv:2110.08454}.

\bibitem[{Lee et~al.(2019)Lee, Kim, and Park}]{lee-etal-2019-domain}
Seanie Lee, Donggyu Kim, and Jangwon Park. 2019.
\newblock Domain-agnostic question-answering with adversarial training.
\newblock In \emph{Proceedings of the 2nd Workshop on Machine Reading for
  Question Answering}. Association for Computational Linguistics.

\bibitem[{Lester et~al.(2021)Lester, Al-Rfou, and Constant}]{lester2021power}
Brian Lester, Rami Al-Rfou, and Noah Constant. 2021.
\newblock The power of scale for parameter-efficient prompt tuning.
\newblock In \emph{Proceedings of the 2021 Conference on Empirical Methods in
  Natural Language Processing}, pages 3045--3059.

\bibitem[{Levy et~al.(2017)Levy, Seo, Choi, and Zettlemoyer}]{levy2017zero}
Omer Levy, Minjoon Seo, Eunsol Choi, and Luke Zettlemoyer. 2017.
\newblock Zero-shot relation extraction via reading comprehension.
\newblock In \emph{Proceedings of the 21st Conference on Computational Natural
  Language Learning (CoNLL 2017)}, pages 333--342.

\bibitem[{Lewis et~al.(2020)Lewis, Oguz, Rinott, Riedel, and
  Schwenk}]{lewis2020mlqa}
Patrick Lewis, Barlas Oguz, Ruty Rinott, Sebastian Riedel, and Holger Schwenk.
  2020.
\newblock Mlqa: Evaluating cross-lingual extractive question answering.
\newblock In \emph{Proceedings of the 58th Annual Meeting of the Association
  for Computational Linguistics}, pages 7315--7330.

\bibitem[{Lewis et~al.(2021)Lewis, Stenetorp, and Riedel}]{lewis2021question}
Patrick Lewis, Pontus Stenetorp, and Sebastian Riedel. 2021.
\newblock Question and answer test-train overlap in open-domain question
  answering datasets.
\newblock In \emph{Proceedings of the 16th Conference of the European Chapter
  of the Association for Computational Linguistics: Main Volume}, pages
  1000--1008.

\bibitem[{Li et~al.(2022)Li, Hou, and Che}]{li2022data}
Bohan Li, Yutai Hou, and Wanxiang Che. 2022.
\newblock Data augmentation approaches in natural language processing: A
  survey.
\newblock \emph{AI Open}.

\bibitem[{Li and Liang(2021)}]{Li2021PrefixTuningOC}
Xiang~Lisa Li and Percy Liang. 2021.
\newblock Prefix-tuning: Optimizing continuous prompts for generation.
\newblock \emph{Proceedings of the 59th Annual Meeting of the Association for
  Computational Linguistics and the 11th International Joint Conference on
  Natural Language Processing (Volume 1: Long Papers)}, abs/2101.00190.

\bibitem[{Li et~al.(2021)Li, Yin, Chen, and Zhang}]{li2021compositional}
Yafu Li, Yongjing Yin, Yulong Chen, and Yue Zhang. 2021.
\newblock On compositional generalization of neural machine translation.
\newblock In \emph{Proceedings of the 59th Annual Meeting of the Association
  for Computational Linguistics and the 11th International Joint Conference on
  Natural Language Processing (Volume 1: Long Papers)}, pages 4767--4780.

\bibitem[{Li et~al.(2019{\natexlab{a}})Li, Zhao, Wang, and
  Hestness}]{li2019compositional}
Yuanpeng Li, Liang Zhao, Jianyu Wang, and Joel Hestness. 2019{\natexlab{a}}.
\newblock Compositional generalization for primitive substitutions.
\newblock In \emph{Proceedings of the 2019 Conference on Empirical Methods in
  Natural Language Processing and the 9th International Joint Conference on
  Natural Language Processing (EMNLP-IJCNLP)}, pages 4293--4302.

\bibitem[{Li et~al.(2019{\natexlab{b}})Li, Li, Wei, Bing, Zhang, and
  Yang}]{li2019transferable}
Zheng Li, Xin Li, Ying Wei, Lidong Bing, Yu~Zhang, and Qiang Yang.
  2019{\natexlab{b}}.
\newblock Transferable end-to-end aspect-based sentiment analysis with
  selective adversarial learning.
\newblock In \emph{Proceedings of the 2019 Conference on Empirical Methods in
  Natural Language Processing and the 9th International Joint Conference on
  Natural Language Processing (EMNLP-IJCNLP)}, pages 4590--4600.

\bibitem[{Lim et~al.(2020)Lim, Lee, Carbonell, and Poibeau}]{lim2020semi}
KyungTae Lim, Jay~Yoon Lee, Jaime Carbonell, and Thierry Poibeau. 2020.
\newblock Semi-supervised learning on meta structure: Multi-task tagging and
  parsing in low-resource scenarios.
\newblock In \emph{Proceedings of the AAAI Conference on Artificial
  Intelligence}, volume~34, pages 8344--8351.

\bibitem[{Liu et~al.(2021{\natexlab{a}})Liu, Niehues, Cross, Guzm{\'a}n, and
  Li}]{liu2021improving}
Danni Liu, Jan Niehues, James Cross, Francisco Guzm{\'a}n, and Xian Li.
  2021{\natexlab{a}}.
\newblock Improving zero-shot translation by disentangling positional
  information.
\newblock In \emph{Proceedings of the 59th Annual Meeting of the Association
  for Computational Linguistics and the 11th International Joint Conference on
  Natural Language Processing (Volume 1: Long Papers)}, pages 1259--1273.

\bibitem[{Liu et~al.(2021{\natexlab{b}})Liu, Haghgoo, Chen, Raghunathan, Koh,
  Sagawa, Liang, and Finn}]{liu2021just}
Evan~Z Liu, Behzad Haghgoo, Annie~S Chen, Aditi Raghunathan, Pang~Wei Koh,
  Shiori Sagawa, Percy Liang, and Chelsea Finn. 2021{\natexlab{b}}.
\newblock Just train twice: Improving group robustness without training group
  information.
\newblock In \emph{International Conference on Machine Learning}, pages
  6781--6792. PMLR.

\bibitem[{Liu et~al.(2021{\natexlab{c}})Liu, Cui, Liu, Huang, Wang, and
  Zhang}]{liu2021logiqa}
Jian Liu, Leyang Cui, Hanmeng Liu, Dandan Huang, Yile Wang, and Yue Zhang.
  2021{\natexlab{c}}.
\newblock Logiqa: a challenge dataset for machine reading comprehension with
  logical reasoning.
\newblock In \emph{Proceedings of the Twenty-Ninth International Conference on
  International Joint Conferences on Artificial Intelligence}, pages
  3622--3628.

\bibitem[{Liu et~al.(2021{\natexlab{d}})Liu, Fu, Tan, Chen, Zhang, Huang, and
  Gao}]{liu2021noisy}
Kun Liu, Yao Fu, Chuanqi Tan, Mosha Chen, Ningyu Zhang, Songfang Huang, and
  Sheng Gao. 2021{\natexlab{d}}.
\newblock Noisy-labeled ner with confidence estimation.
\newblock In \emph{Proceedings of the 2021 Conference of the North American
  Chapter of the Association for Computational Linguistics: Human Language
  Technologies}, pages 3437--3445.

\bibitem[{Liu et~al.(2021{\natexlab{e}})Liu, Lewis, Riedel, and
  Stenetorp}]{liu2021challenges}
Linqing Liu, Patrick Lewis, Sebastian Riedel, and Pontus Stenetorp.
  2021{\natexlab{e}}.
\newblock Challenges in generalization in open domain question answering.
\newblock \emph{arXiv preprint arXiv:2109.01156}.

\bibitem[{Liu et~al.(2019{\natexlab{a}})Liu, Gardner, Belinkov, Peters, and
  Smith}]{liu-gardner-belinkov-peters-smith:2019:NAACL}
Nelson~F. Liu, Matt Gardner, Yonatan Belinkov, Matthew~E. Peters, and Noah~A.
  Smith. 2019{\natexlab{a}}.
\newblock Linguistic knowledge and transferability of contextual
  representations.
\newblock In \emph{Proceedings of the Conference of the North American Chapter
  of the Association for Computational Linguistics: Human Language
  Technologies}.

\bibitem[{Liu et~al.(2021{\natexlab{f}})Liu, Yuan, Fu, Jiang, Hayashi, and
  Neubig}]{liu2021pre}
Pengfei Liu, Weizhe Yuan, Jinlan Fu, Zhengbao Jiang, Hiroaki Hayashi, and
  Graham Neubig. 2021{\natexlab{f}}.
\newblock Pre-train, prompt, and predict: A systematic survey of prompting
  methods in natural language processing.
\newblock \emph{arXiv preprint arXiv:2107.13586}.

\bibitem[{Liu et~al.(2018)Liu, Zhang, and Liu}]{liu2018learning}
Qi~Liu, Yue Zhang, and Jiangming Liu. 2018.
\newblock Learning domain representation for multi-domain sentiment
  classification.
\newblock In \emph{Proceedings of the 2018 Conference of the North American
  Chapter of the Association for Computational Linguistics: Human Language
  Technologies, Volume 1 (Long Papers)}, pages 541--550.

\bibitem[{Liu et~al.(2020)Liu, Xin, Chang, and Sui}]{liu-etal-2020-hyponli}
Tianyu Liu, Zheng Xin, Baobao Chang, and Zhifang Sui. 2020.
\newblock \href {https://aclanthology.org/2020.lrec-1.846} {{H}ypo{NLI}:
  Exploring the artificial patterns of hypothesis-only bias in natural language
  inference}.
\newblock In \emph{Proceedings of the 12th Language Resources and Evaluation
  Conference}, pages 6852--6860, Marseille, France. European Language Resources
  Association.

\bibitem[{Liu et~al.(2021{\natexlab{g}})Liu, Ji, Fu, Du, Yang, and
  Tang}]{liu2021p}
Xiao Liu, Kaixuan Ji, Yicheng Fu, Zhengxiao Du, Zhilin Yang, and Jie Tang.
  2021{\natexlab{g}}.
\newblock P-tuning v2: Prompt tuning can be comparable to fine-tuning
  universally across scales and tasks.
\newblock \emph{arXiv preprint arXiv:2110.07602}.

\bibitem[{Liu et~al.(2019{\natexlab{b}})Liu, Ott, Goyal, Du, Joshi, Chen, Levy,
  Lewis, Zettlemoyer, and Stoyanov}]{liu2019roberta}
Yinhan Liu, Myle Ott, Naman Goyal, Jingfei Du, Mandar Joshi, Danqi Chen, Omer
  Levy, Mike Lewis, Luke Zettlemoyer, and Veselin Stoyanov. 2019{\natexlab{b}}.
\newblock Roberta: A robustly optimized bert pretraining approach.
\newblock \emph{arXiv preprint arXiv:1907.11692}.

\bibitem[{Liu et~al.(2021{\natexlab{h}})Liu, Xu, Yu, Dai, Ji, Cahyawijaya,
  Madotto, and Fung}]{liu2021crossner}
Zihan Liu, Yan Xu, Tiezheng Yu, Wenliang Dai, Ziwei Ji, Samuel Cahyawijaya,
  Andrea Madotto, and Pascale Fung. 2021{\natexlab{h}}.
\newblock Crossner: Evaluating cross-domain named entity recognition.
\newblock In \emph{Proceedings of the AAAI Conference on Artificial
  Intelligence}, volume~35, pages 13452--13460.

\bibitem[{Lu et~al.(2022)Lu, Yang, Mac~Namee, and Zhang}]{lu2022rationale}
Jinghui Lu, Linyi Yang, Brian Mac~Namee, and Yue Zhang. 2022.
\newblock A rationale-centric framework for human-in-the-loop machine learning.
\newblock \emph{arXiv preprint arXiv:2203.12918}.

\bibitem[{Lu et~al.(2020)Lu, Mardziel, Wu, Amancharla, and
  Datta}]{lu2020gender}
Kaiji Lu, Piotr Mardziel, Fangjing Wu, Preetam Amancharla, and Anupam Datta.
  2020.
\newblock Gender bias in neural natural language processing.
\newblock In \emph{Logic, Language, and Security}, pages 189--202. Springer.

\bibitem[{Lyu et~al.(2022)Lyu, Foster, and Graham}]{lyu2022extending}
Chenyang Lyu, Jennifer Foster, and Yvette Graham. 2022.
\newblock Extending the scope of out-of-domain: Examining qa models in multiple
  subdomains.
\newblock \emph{arXiv preprint arXiv:2204.04534}.

\bibitem[{Ma et~al.(2021)Ma, Zhou, Gui, Tan, Zhang, and Huang}]{ma2021template}
Ruotian Ma, Xin Zhou, Tao Gui, Yiding Tan, Qi~Zhang, and Xuanjing Huang. 2021.
\newblock Template-free prompt tuning for few-shot ner.
\newblock \emph{arXiv preprint arXiv:2109.13532}.

\bibitem[{Maudslay et~al.(2019)Maudslay, Gonen, Cotterell, and
  Teufel}]{maudslay2019s}
Rowan~Hall Maudslay, Hila Gonen, Ryan Cotterell, and Simone Teufel. 2019.
\newblock It’s all in the name: Mitigating gender bias with name-based
  counterfactual data substitution.
\newblock In \emph{Proceedings of the 2019 Conference on Empirical Methods in
  Natural Language Processing and the 9th International Joint Conference on
  Natural Language Processing (EMNLP-IJCNLP)}, pages 5270--5278.

\bibitem[{May et~al.(2019)May, Wang, Bordia, Bowman, and
  Rudinger}]{may2019measuring}
Chandler May, Alex Wang, Shikha Bordia, Samuel Bowman, and Rachel Rudinger.
  2019.
\newblock On measuring social biases in sentence encoders.
\newblock In \emph{Proceedings of the 2019 Conference of the North American
  Chapter of the Association for Computational Linguistics: Human Language
  Technologies, Volume 1 (Long and Short Papers)}, pages 622--628.

\bibitem[{McCoy et~al.(2019)McCoy, Pavlick, and Linzen}]{mccoy2019right}
Tom McCoy, Ellie Pavlick, and Tal Linzen. 2019.
\newblock Right for the wrong reasons: Diagnosing syntactic heuristics in
  natural language inference.
\newblock In \emph{Proceedings of the 57th Annual Meeting of the Association
  for Computational Linguistics}, pages 3428--3448.

\bibitem[{Miao et~al.(2020)Miao, Li, Wang, and Tan}]{miao2020snippext}
Zhengjie Miao, Yuliang Li, Xiaolan Wang, and Wang-Chiew Tan. 2020.
\newblock Snippext: Semi-supervised opinion mining with augmented data.
\newblock In \emph{Proceedings of The Web Conference 2020}, pages 617--628.

\bibitem[{Michel et~al.(2020)Michel, Hashimoto, and
  Neubig}]{michel2020modeling}
Paul Michel, Tatsunori Hashimoto, and Graham Neubig. 2020.
\newblock Modeling the second player in distributionally robust optimization.
\newblock In \emph{International Conference on Learning Representations}.

\bibitem[{Min et~al.(2019)Min, Wallace, Singh, Gardner, Hajishirzi, and
  Zettlemoyer}]{min2019compositional}
Sewon Min, Eric Wallace, Sameer Singh, Matt Gardner, Hannaneh Hajishirzi, and
  Luke Zettlemoyer. 2019.
\newblock Compositional questions do not necessitate multi-hop reasoning.
\newblock In \emph{Proceedings of the 57th Annual Meeting of the Association
  for Computational Linguistics}, pages 4249--4257.

\bibitem[{Miyato et~al.(2017)Miyato, Dai, and
  Goodfellow}]{Miyato2017AdversarialTM}
Takeru Miyato, Andrew~M. Dai, and Ian~J. Goodfellow. 2017.
\newblock Adversarial training methods for semi-supervised text classification.
\newblock \emph{arXiv: Machine Learning}.

\bibitem[{Moradi et~al.(2021)Moradi, Blagec, and Samwald}]{moradi2021deep}
Milad Moradi, Kathrin Blagec, and Matthias Samwald. 2021.
\newblock Deep learning models are not robust against noise in clinical text.
\newblock \emph{arXiv preprint arXiv:2108.12242}.

\bibitem[{Moradi and Samwald(2021)}]{moradi2021evaluating}
Milad Moradi and Matthias Samwald. 2021.
\newblock Evaluating the robustness of neural language models to input
  perturbations.
\newblock In \emph{Proceedings of the 2021 Conference on Empirical Methods in
  Natural Language Processing}, pages 1558--1570.

\bibitem[{Morgan and Winship(2015)}]{morgan2015counterfactuals}
Stephen~L Morgan and Christopher Winship. 2015.
\newblock \emph{Counterfactuals and causal inference}.
\newblock Cambridge University Press.

\bibitem[{Naik et~al.(2018)Naik, Ravichander, Sadeh, Rose, and
  Neubig}]{naik2018stress}
Aakanksha Naik, Abhilasha Ravichander, Norman Sadeh, Carolyn Rose, and Graham
  Neubig. 2018.
\newblock Stress test evaluation for natural language inference.
\newblock In \emph{Proceedings of the 27th International Conference on
  Computational Linguistics}, pages 2340--2353.

\bibitem[{N{\'a}plava et~al.(2021)N{\'a}plava, Popel, Straka, and
  Strakov{\'a}}]{naplava2021understanding}
Jakub N{\'a}plava, Martin Popel, Milan Straka, and Jana Strakov{\'a}. 2021.
\newblock Understanding model robustness to user-generated noisy texts.
\newblock In \emph{Proceedings of the Seventh Workshop on Noisy User-generated
  Text (W-NUT 2021)}, pages 340--350.

\bibitem[{Nguyen et~al.(2021)Nguyen, Gelli, and Poria}]{Nguyen2021DOZENCZ}
Hoang Nguyen, Francesco Gelli, and Soujanya Poria. 2021.
\newblock Dozen: Cross-domain zero shot named entity recognition with knowledge
  graph.
\newblock \emph{Proceedings of the 44th International ACM SIGIR Conference on
  Research and Development in Information Retrieval}.

\bibitem[{Ni et~al.(2019)Ni, Li, and McAuley}]{ni2019justifying}
Jianmo Ni, Jiacheng Li, and Julian McAuley. 2019.
\newblock Justifying recommendations using distantly-labeled reviews and
  fine-grained aspects.
\newblock In \emph{Proceedings of the 2019 Conference on Empirical Methods in
  Natural Language Processing and the 9th International Joint Conference on
  Natural Language Processing (EMNLP-IJCNLP)}, pages 188--197.

\bibitem[{Nie et~al.(2020)Nie, Williams, Dinan, Bansal, Weston, and
  Kiela}]{nie-etal-2020-adversarial}
Yixin Nie, Adina Williams, Emily Dinan, Mohit Bansal, Jason Weston, and Douwe
  Kiela. 2020.
\newblock \href {https://doi.org/10.18653/v1/2020.acl-main.441} {Adversarial
  {NLI}: A new benchmark for natural language understanding}.
\newblock In \emph{Proceedings of the 58th Annual Meeting of the Association
  for Computational Linguistics}, pages 4885--4901, Online. Association for
  Computational Linguistics.

\bibitem[{Niu et~al.(2023)Niu, Yang, Dong, and Zhang}]{niu2023learning}
Yingjie Niu, Linyi Yang, Ruihai Dong, and Yue Zhang. 2023.
\newblock Learning to generalize for cross-domain qa.
\newblock \emph{arXiv preprint arXiv:2305.08208}.

\bibitem[{Oren et~al.(2019)Oren, Sagawa, Hashimoto, and
  Liang}]{oren2019distributionally}
Yonatan Oren, Shiori Sagawa, Tatsunori~B Hashimoto, and Percy Liang. 2019.
\newblock Distributionally robust language modeling.
\newblock In \emph{Proceedings of the 2019 Conference on Empirical Methods in
  Natural Language Processing and the 9th International Joint Conference on
  Natural Language Processing (EMNLP-IJCNLP)}, pages 4227--4237.

\bibitem[{Pan et~al.(2010)Pan, Ni, Sun, Yang, and Chen}]{pan2010cross}
Sinno~Jialin Pan, Xiaochuan Ni, Jian-Tao Sun, Qiang Yang, and Zheng Chen. 2010.
\newblock Cross-domain sentiment classification via spectral feature alignment.
\newblock In \emph{Proceedings of the 19th international conference on World
  wide web}, pages 751--760.

\bibitem[{Park et~al.(2018)Park, Shin, and Fung}]{park2018reducing}
Ji~Ho Park, Jamin Shin, and Pascale Fung. 2018.
\newblock Reducing gender bias in abusive language detection.
\newblock In \emph{Proceedings of the 2018 Conference on Empirical Methods in
  Natural Language Processing}, pages 2799--2804.

\bibitem[{Pearl et~al.(2000)}]{pearl2000models}
Judea Pearl et~al. 2000.
\newblock Models, reasoning and inference.
\newblock \emph{Cambridge, UK: CambridgeUniversityPress}, 19:2.

\bibitem[{Petroni et~al.(2019)Petroni, Rockt{\"a}schel, Lewis, Bakhtin, Wu,
  Miller, and Riedel}]{petroni2019language}
Fabio Petroni, Tim Rockt{\"a}schel, Patrick Lewis, Anton Bakhtin, Yuxiang Wu,
  Alexander~H Miller, and Sebastian Riedel. 2019.
\newblock Language models as knowledge bases?
\newblock \emph{arXiv preprint arXiv:1909.01066}.

\bibitem[{Peyrard et~al.(2021)Peyrard, Ghotra, Josifoski, Agarwal, Patra,
  Carignan, Kiciman, and West}]{peyrard2021invariant}
Maxime Peyrard, Sarvjeet~Singh Ghotra, Martin Josifoski, Vidhan Agarwal, Barun
  Patra, Dean Carignan, Emre Kiciman, and Robert West. 2021.
\newblock Invariant language modeling.
\newblock \emph{arXiv preprint arXiv:2110.08413}.

\bibitem[{Pezeshkpour et~al.(2022)Pezeshkpour, Jain, Singh, and
  Wallace}]{pezeshkpour2022combining}
Pouya Pezeshkpour, Sarthak Jain, Sameer Singh, and Byron~C Wallace. 2022.
\newblock Combining feature and instance attribution to detect artifacts.
\newblock In \emph{Proceedings of the 60th Annual Meeting of the Association
  for Computational Linguistics (ACL 2022)}, pages 5533--5538.

\bibitem[{Pfeiffer et~al.(2020)Pfeiffer, Vuli{\'c}, Gurevych, and
  Ruder}]{pfeiffer-etal-2020-mad}
Jonas Pfeiffer, Ivan Vuli{\'c}, Iryna Gurevych, and Sebastian Ruder. 2020.
\newblock {MAD-X}: {A}n {A}dapter-{B}ased {F}ramework for {M}ulti-{T}ask
  {C}ross-{L}ingual {T}ransfer.
\newblock In \emph{Proceedings of the 2020 Conference on Empirical Methods in
  Natural Language Processing (EMNLP)}. Association for Computational
  Linguistics.

\bibitem[{Pichai(2023)}]{pichai2023important}
Sundar Pichai. 2023.
\newblock An important next step on our ai journey.
\newblock \emph{Google Blog, The Keyword}.

\bibitem[{Plank(2021)}]{plank2021cross}
Barbara Plank. 2021.
\newblock Cross-lingual cross-domain nested named entity evaluation on english
  web texts.
\newblock In \emph{Findings of the Association for Computational Linguistics:
  ACL-IJCNLP 2021}, pages 1808--1815.

\bibitem[{Poliak et~al.(2018)Poliak, Naradowsky, Haldar, Rudinger, and
  Van~Durme}]{poliak2018hypothesis}
Adam Poliak, Jason Naradowsky, Aparajita Haldar, Rachel Rudinger, and Benjamin
  Van~Durme. 2018.
\newblock Hypothesis only baselines in natural language inference.
\newblock In \emph{Proceedings of the Seventh Joint Conference on Lexical and
  Computational Semantics}, pages 180--191.

\bibitem[{Pryzant et~al.(2018)Pryzant, Shen, Jurafsky, and
  Wagner}]{pryzant2018deconfounded}
Reid Pryzant, Kelly Shen, Dan Jurafsky, and Stefan Wagner. 2018.
\newblock Deconfounded lexicon induction for interpretable social science.
\newblock In \emph{Proceedings of the 2018 Conference of the North American
  Chapter of the Association for Computational Linguistics: Human Language
  Technologies, Volume 1 (Long Papers)}, pages 1615--1625.

\bibitem[{Qin and Eisner(2021)}]{qin-eisner-2021-learning}
Guanghui Qin and Jason Eisner. 2021.
\newblock Learning how to ask: Querying {LM}s with mixtures of soft prompts.
\newblock In \emph{Proceedings of the 2021 Conference of the North American
  Chapter of the Association for Computational Linguistics: Human Language
  Technologies}. Association for Computational Linguistics.

\bibitem[{Quionero-Candela et~al.(2009)Quionero-Candela, Sugiyama,
  Schwaighofer, and Lawrence}]{quionero2009dataset}
Joaquin Quionero-Candela, Masashi Sugiyama, Anton Schwaighofer, and Neil~D
  Lawrence. 2009.
\newblock Dataset shift in machine learning.

\bibitem[{Radford et~al.(2018)Radford, Narasimhan, Salimans, and
  Sutskever}]{radford2018improving}
Alec Radford, Karthik Narasimhan, Tim Salimans, and Ilya Sutskever. 2018.
\newblock Improving language understanding by generative pre-training.

\bibitem[{Radford et~al.(2019)Radford, Wu, Child, Luan, Amodei, and
  Sutskever}]{Radford2019LanguageMA}
Alec Radford, Jeff Wu, Rewon Child, David Luan, Dario Amodei, and Ilya
  Sutskever. 2019.
\newblock Language models are unsupervised multitask learners.

\bibitem[{Ramponi and Plank(2020)}]{ramponi2020neural}
Alan Ramponi and Barbara Plank. 2020.
\newblock Neural unsupervised domain adaptation in nlp—a survey.
\newblock In \emph{Proceedings of the 28th International Conference on
  Computational Linguistics}, pages 6838--6855.

\bibitem[{Rebuffi et~al.(2017)Rebuffi, Bilen, and
  Vedaldi}]{rebuffi2017learning}
Sylvestre-Alvise Rebuffi, Hakan Bilen, and Andrea Vedaldi. 2017.
\newblock Learning multiple visual domains with residual adapters.
\newblock \emph{Advances in neural information processing systems}, 30.

\bibitem[{Ribeiro et~al.(2020)Ribeiro, Wu, Guestrin, and
  Singh}]{ribeiro2020beyond}
Marco~Tulio Ribeiro, Tongshuang Wu, Carlos Guestrin, and Sameer Singh. 2020.
\newblock Beyond accuracy: Behavioral testing of nlp models with checklist.
\newblock In \emph{Proceedings of the 58th Annual Meeting of the Association
  for Computational Linguistics}, pages 4902--4912.

\bibitem[{Rogers et~al.(2021)Rogers, Gardner, and Augenstein}]{rogers2021qa}
Anna Rogers, Matt Gardner, and Isabelle Augenstein. 2021.
\newblock Qa dataset explosion: A taxonomy of nlp resources for question
  answering and reading comprehension.
\newblock \emph{arXiv preprint arXiv:2107.12708}.

\bibitem[{Ruder and Plank(2018)}]{ruder2018strong}
Sebastian Ruder and Barbara Plank. 2018.
\newblock Strong baselines for neural semi-supervised learning under domain
  shift.
\newblock In \emph{Proceedings of the 56th Annual Meeting of the Association
  for Computational Linguistics (Volume 1: Long Papers)}, pages 1044--1054.

\bibitem[{Rudinger et~al.(2017)Rudinger, May, and
  Van~Durme}]{rudinger2017social}
Rachel Rudinger, Chandler May, and Benjamin Van~Durme. 2017.
\newblock Social bias in elicited natural language inferences.
\newblock In \emph{Proceedings of the First ACL Workshop on Ethics in Natural
  Language Processing}, pages 74--79.

\bibitem[{Rudinger et~al.(2018)Rudinger, Naradowsky, Leonard, and
  Van~Durme}]{rudinger-etal-2018-gender}
Rachel Rudinger, Jason Naradowsky, Brian Leonard, and Benjamin Van~Durme. 2018.
\newblock \href {https://doi.org/10.18653/v1/N18-2002} {Gender bias in
  coreference resolution}.
\newblock In \emph{Proceedings of the 2018 Conference of the North {A}merican
  Chapter of the Association for Computational Linguistics: Human Language
  Technologies, Volume 2 (Short Papers)}, pages 8--14, New Orleans, Louisiana.
  Association for Computational Linguistics.

\bibitem[{Russin et~al.(2019)Russin, Jo, O'Reilly, and
  Bengio}]{russin2019compositional}
Jake Russin, Jason Jo, Randall~C O'Reilly, and Yoshua Bengio. 2019.
\newblock Compositional generalization in a deep seq2seq model by separating
  syntax and semantics.
\newblock \emph{arXiv preprint arXiv:1904.09708}.

\bibitem[{Rychalska et~al.(2019)Rychalska, Basaj, Gosiewska, and
  Biecek}]{rychalska2019models}
Barbara Rychalska, Dominika Basaj, Alicja Gosiewska, and Przemys{\l}aw Biecek.
  2019.
\newblock Models in the wild: On corruption robustness of neural nlp systems.
\newblock In \emph{International Conference on Neural Information Processing},
  pages 235--247. Springer.

\bibitem[{Sagawa et~al.(2020)Sagawa, Koh, Hashimoto, and
  Liang}]{sagawa2019distributionally}
Shiori Sagawa, Pang~Wei Koh, Tatsunori~B Hashimoto, and Percy Liang. 2020.
\newblock Distributionally robust neural networks for group shifts: On the
  importance of regularization for worst-case generalization.
\newblock In \emph{International Conference on Learning Representations}.

\bibitem[{Saha et~al.(2019)Saha, Sugar, Torous, Abrahao, Kıcıman, and
  Choudhury}]{Saha2019ASM}
Koustuv Saha, Benjamin Sugar, John~B Torous, Bruno~D. Abrahao, Emre Kıcıman,
  and Munmun~De Choudhury. 2019.
\newblock A social media study on the effects of psychiatric medication use.
\newblock \emph{Proceedings of the ... International AAAI Conference on Weblogs
  and Social Media. International AAAI Conference on Weblogs and Social Media},
  13:440--451.

\bibitem[{Sakaguchi et~al.(2020)Sakaguchi, Le~Bras, Bhagavatula, and
  Choi}]{sakaguchi2020winogrande}
Keisuke Sakaguchi, Ronan Le~Bras, Chandra Bhagavatula, and Yejin Choi. 2020.
\newblock Winogrande: An adversarial winograd schema challenge at scale.
\newblock In \emph{Proceedings of the AAAI Conference on Artificial
  Intelligence}, volume~34, pages 8732--8740.

\bibitem[{Sang and De~Meulder(2003)}]{sang2003introduction}
Erik Tjong~Kim Sang and Fien De~Meulder. 2003.
\newblock Introduction to the conll-2003 shared task: Language-independent
  named entity recognition.
\newblock In \emph{Proceedings of the Seventh Conference on Natural Language
  Learning at HLT-NAACL 2003}, pages 142--147.

\bibitem[{Schlegel et~al.(2020)Schlegel, Nenadic, and
  Batista-Navarro}]{schlegel2020beyond}
Viktor Schlegel, Goran Nenadic, and Riza Batista-Navarro. 2020.
\newblock Beyond leaderboards: A survey of methods for revealing weaknesses in
  natural language inference data and models.
\newblock \emph{arXiv preprint arXiv:2005.14709}.

\bibitem[{Sch{\"o}lkopf et~al.(2021)Sch{\"o}lkopf, Locatello, Bauer, Ke,
  Kalchbrenner, Goyal, and Bengio}]{scholkopf2021toward}
Bernhard Sch{\"o}lkopf, Francesco Locatello, Stefan Bauer, Nan~Rosemary Ke, Nal
  Kalchbrenner, Anirudh Goyal, and Yoshua Bengio. 2021.
\newblock Toward causal representation learning.
\newblock \emph{Proceedings of the IEEE}, 109(5):612--634.

\bibitem[{Shen et~al.(2021)Shen, Liu, He, Zhang, Xu, Yu, and Cui}]{ood}
Zheyan Shen, Jiashuo Liu, Yue He, Xingxuan Zhang, Renzhe Xu, Han Yu, and Peng
  Cui. 2021.
\newblock \href {http://arxiv.org/abs/2108.13624} {Towards out-of-distribution
  generalization: {A} survey}.
\newblock \emph{CoRR}, abs/2108.13624.

\bibitem[{Snell et~al.(2017)Snell, Swersky, and Zemel}]{snell2017prototypical}
Jake Snell, Kevin Swersky, and Richard Zemel. 2017.
\newblock Prototypical networks for few-shot learning.
\newblock \emph{Advances in neural information processing systems}, 30.

\bibitem[{Srivastava et~al.(2020)Srivastava, Hashimoto, and
  Liang}]{srivastava2020robustness}
Megha Srivastava, Tatsunori Hashimoto, and Percy Liang. 2020.
\newblock Robustness to spurious correlations via human annotations.
\newblock In \emph{International Conference on Machine Learning}, pages
  9109--9119. PMLR.

\bibitem[{Stanovsky et~al.(2019)Stanovsky, Smith, and
  Zettlemoyer}]{stanovsky2019evaluating}
Gabriel Stanovsky, Noah~A Smith, and Luke Zettlemoyer. 2019.
\newblock Evaluating gender bias in machine translation.
\newblock In \emph{Proceedings of the 57th Annual Meeting of the Association
  for Computational Linguistics}, pages 1679--1684.

\bibitem[{Sugawara et~al.(2018)Sugawara, Inui, Sekine, and
  Aizawa}]{sugawara2018makes}
Saku Sugawara, Kentaro Inui, Satoshi Sekine, and Akiko Aizawa. 2018.
\newblock What makes reading comprehension questions easier?
\newblock In \emph{Proceedings of the 2018 Conference on Empirical Methods in
  Natural Language Processing}, pages 4208--4219.

\bibitem[{Sugawara et~al.(2020)Sugawara, Stenetorp, Inui, and
  Aizawa}]{sugawara2020assessing}
Saku Sugawara, Pontus Stenetorp, Kentaro Inui, and Akiko Aizawa. 2020.
\newblock Assessing the benchmarking capacity of machine reading comprehension
  datasets.
\newblock In \emph{Proceedings of the AAAI Conference on Artificial
  Intelligence}, volume~34, pages 8918--8927.

\bibitem[{Tarcar et~al.(2019)Tarcar, Tiwari, Dhaimodker, Rebelo, Desai, and
  Rao}]{tarcar2019healthcare}
Amogh~Kamat Tarcar, Aashis Tiwari, Vineet~Naique Dhaimodker, Penjo Rebelo,
  Rahul Desai, and Dattaraj Rao. 2019.
\newblock Healthcare ner models using language model pretraining.
\newblock \emph{arXiv preprint arXiv:1910.11241}.

\bibitem[{Tateisi et~al.(2005)Tateisi, Yakushiji, Ohta, and
  Tsujii}]{tateisi-etal-2005-syntax}
Yuka Tateisi, Akane Yakushiji, Tomoko Ohta, and Jun{'}ichi Tsujii. 2005.
\newblock \href {https://aclanthology.org/I05-2038} {Syntax annotation for the
  {GENIA} corpus}.
\newblock In \emph{Companion Volume to the Proceedings of Conference including
  Posters/Demos and tutorial abstracts}.

\bibitem[{Tomalin et~al.(2021)Tomalin, Byrne, Concannon, Saunders, and
  Ullmann}]{tomalin2021practical}
Marcus Tomalin, Bill Byrne, Shauna Concannon, Danielle Saunders, and Stefanie
  Ullmann. 2021.
\newblock The practical ethics of bias reduction in machine translation: why
  domain adaptation is better than data debiasing.
\newblock \emph{Ethics and Information Technology}, 23(3):419--433.

\bibitem[{Touvron et~al.(2023)Touvron, Lavril, Izacard, Martinet, Lachaux,
  Lacroix, Rozi{\`e}re, Goyal, Hambro, Azhar et~al.}]{touvron2023llama}
Hugo Touvron, Thibaut Lavril, Gautier Izacard, Xavier Martinet, Marie-Anne
  Lachaux, Timoth{\'e}e Lacroix, Baptiste Rozi{\`e}re, Naman Goyal, Eric
  Hambro, Faisal Azhar, et~al. 2023.
\newblock Llama: Open and efficient foundation language models.
\newblock \emph{arXiv preprint arXiv:2302.13971}.

\bibitem[{Tu et~al.(2020)Tu, Lalwani, Gella, and He}]{tu2020empirical}
Lifu Tu, Garima Lalwani, Spandana Gella, and He~He. 2020.
\newblock An empirical study on robustness to spurious correlations using
  pre-trained language models.
\newblock \emph{Transactions of the Association for Computational Linguistics},
  8:621--633.

\bibitem[{Utama et~al.(2021)Utama, Moosavi, Sanh, and
  Gurevych}]{utama2021avoiding}
Prasetya~Ajie Utama, Nafise~Sadat Moosavi, Victor Sanh, and Iryna Gurevych.
  2021.
\newblock Avoiding inference heuristics in few-shot prompt-based finetuning.
\newblock \emph{Proceedings of the Conference on Empirical Methods in Natural
  Language Processing (EMNLP)}.

\bibitem[{Vanmassenhove et~al.(2018)Vanmassenhove, Hardmeier, and
  Way}]{vanmassenhove2018getting}
Eva Vanmassenhove, Christian Hardmeier, and Andy Way. 2018.
\newblock Getting gender right in neural machine translation.
\newblock In \emph{Proceedings of the 2018 Conference on Empirical Methods in
  Natural Language Processing}, pages 3003--3008.

\bibitem[{Varshney et~al.(2020)Varshney, Mishra, and
  Baral}]{Varshney2020TowardsIS}
Neeraj Varshney, Swaroop Mishra, and Chitta Baral. 2020.
\newblock Towards improving selective prediction ability of nlp systems.

\bibitem[{Veitch et~al.(2021)Veitch, D'Amour, Yadlowsky, and
  Eisenstein}]{veitch2021counterfactual}
Victor Veitch, Alexander D'Amour, Steve Yadlowsky, and Jacob Eisenstein. 2021.
\newblock Counterfactual invariance to spurious correlations: Why and how to
  pass stress tests.
\newblock \emph{arXiv preprint arXiv:2106.00545}.

\bibitem[{Vernikos et~al.(2020)Vernikos, Margatina, Chronopoulou, and
  Androutsopoulos}]{vernikos2020domain}
Giorgos Vernikos, Katerina Margatina, Alexandra Chronopoulou, and Ion
  Androutsopoulos. 2020.
\newblock Domain adversarial fine-tuning as an effective regularizer.
\newblock In \emph{Findings of the Association for Computational Linguistics:
  EMNLP 2020}, pages 3103--3112.

\bibitem[{Vu et~al.(2020)Vu, Phung, and Haffari}]{vu2020effective}
Thuy Vu, Dinh Phung, and Gholamreza Haffari. 2020.
\newblock Effective unsupervised domain adaptation with adversarially trained
  language models.
\newblock In \emph{Proceedings of the 2020 Conference on Empirical Methods in
  Natural Language Processing (EMNLP)}, pages 6163--6173.

\bibitem[{Wang et~al.(2018)Wang, Singh, Michael, Hill, Levy, and
  Bowman}]{wang2018glue}
Alex Wang, Amanpreet Singh, Julian Michael, Felix Hill, Omer Levy, and Samuel
  Bowman. 2018.
\newblock Glue: A multi-task benchmark and analysis platform for natural
  language understanding.
\newblock In \emph{Proceedings of the 2018 EMNLP Workshop BlackboxNLP:
  Analyzing and Interpreting Neural Networks for NLP}, pages 353--355.

\bibitem[{Wang et~al.()Wang, Gan, Liu, Liu, Gao, and
  Wang}]{wang-etal-2019-adversarial}
Huazheng Wang, Zhe Gan, Xiaodong Liu, Jingjing Liu, Jianfeng Gao, and Hongning
  Wang.
\newblock Adversarial domain adaptation for machine reading comprehension.
\newblock In \emph{Proceedings of the 2019 Conference on Empirical Methods in
  Natural Language Processing and the 9th International Joint Conference on
  Natural Language Processing (EMNLP-IJCNLP)}. Association for Computational
  Linguistics.

\bibitem[{Wang et~al.(2021{\natexlab{a}})Wang, Yang, and
  Wang}]{wang2021identifying}
Tianlu Wang, Diyi Yang, and Xuezhi Wang. 2021{\natexlab{a}}.
\newblock Identifying and mitigating spurious correlations for improving
  robustness in nlp models.
\newblock \emph{arXiv preprint arXiv:2110.07736}.

\bibitem[{Wang et~al.(2022)Wang, Dou, Xiong, Zou, Zhang, Gui, Qiao, Cheng, and
  Huang}]{wang2022miner}
Xiao Wang, Shihan Dou, Limao Xiong, Yicheng Zou, Qi~Zhang, Tao Gui, Liang Qiao,
  Zhanzhan Cheng, and Xuanjing Huang. 2022.
\newblock Miner: Improving out-of-vocabulary named entity recognition from an
  information theoretic perspective.
\newblock \emph{arXiv preprint arXiv:2204.04391}.

\bibitem[{Wang et~al.(2021{\natexlab{b}})Wang, Liu, Gui, Zhang, Zou, Zhou, Ye,
  Zhang, Zheng, Pang, Wu, Li, Zhang, Ma, Fei, Cai, Zhao, Hu, Yan, Tan, Hu,
  Bian, Liu, Qin, Zhu, Xing, Fu, Zhang, Peng, Zheng, Zhou, Wei, Qiu, and
  Huang}]{wang-etal-2021-textflint}
Xiao Wang, Qin Liu, Tao Gui, Qi~Zhang, Yicheng Zou, Xin Zhou, Jiacheng Ye,
  Yongxin Zhang, Rui Zheng, Zexiong Pang, Qinzhuo Wu, Zhengyan Li, Chong Zhang,
  Ruotian Ma, Zichu Fei, Ruijian Cai, Jun Zhao, Xingwu Hu, Zhiheng Yan, Yiding
  Tan, Yuan Hu, Qiyuan Bian, Zhihua Liu, Shan Qin, Bolin Zhu, Xiaoyu Xing,
  Jinlan Fu, Yue Zhang, Minlong Peng, Xiaoqing Zheng, Yaqian Zhou, Zhongyu Wei,
  Xipeng Qiu, and Xuanjing Huang. 2021{\natexlab{b}}.
\newblock \href {https://doi.org/10.18653/v1/2021.acl-demo.41} {{T}ext{F}lint:
  Unified multilingual robustness evaluation toolkit for natural language
  processing}.
\newblock In \emph{Proceedings of the 59th Annual Meeting of the Association
  for Computational Linguistics and the 11th International Joint Conference on
  Natural Language Processing: System Demonstrations}, pages 347--355, Online.
  Association for Computational Linguistics.

\bibitem[{Wang et~al.(2021{\natexlab{c}})Wang, Wang, and
  Yang}]{wang2021measure}
Xuezhi Wang, Haohan Wang, and Diyi Yang. 2021{\natexlab{c}}.
\newblock Measure and improve robustness in nlp models: A survey.
\newblock \emph{arXiv preprint arXiv:2112.08313}.

\bibitem[{Wang and Culotta(2020)}]{Wang2020IdentifyingSC}
Zhao Wang and Aron Culotta. 2020.
\newblock Identifying spurious correlations for robust text classification.
\newblock In \emph{Findings of the Association for Computational Linguistics:
  EMNLP 2020}, pages 3431--3440.

\bibitem[{Wang and Culotta(2021)}]{wang2021robustness}
Zhao Wang and Aron Culotta. 2021.
\newblock Robustness to spurious correlations in text classification via
  automatically generated counterfactuals.
\newblock In \emph{Proceedings of the AAAI Conference on Artificial
  Intelligence}, volume~35, pages 14024--14031.

\bibitem[{Warstadt et~al.(2020)Warstadt, Parrish, Liu, Mohananey, Peng, Wang,
  and Bowman}]{warstadt2020blimp}
Alex Warstadt, Alicia Parrish, Haokun Liu, Anhad Mohananey, Wei Peng, Sheng-Fu
  Wang, and Samuel~R Bowman. 2020.
\newblock Blimp: The benchmark of linguistic minimal pairs for english.
\newblock \emph{Transactions of the Association for Computational Linguistics},
  8:377--392.

\bibitem[{Wei and Zou(2019)}]{wei2019eda}
Jason Wei and Kai Zou. 2019.
\newblock Eda: Easy data augmentation techniques for boosting performance on
  text classification tasks.
\newblock In \emph{Proceedings of the 2019 Conference on Empirical Methods in
  Natural Language Processing and the 9th International Joint Conference on
  Natural Language Processing (EMNLP-IJCNLP)}, pages 6382--6388.

\bibitem[{Williams et~al.(2018)Williams, Nangia, and
  Bowman}]{williams2018broad}
Adina Williams, Nikita Nangia, and Samuel Bowman. 2018.
\newblock A broad-coverage challenge corpus for sentence understanding through
  inference.
\newblock In \emph{Proceedings of the 2018 Conference of the North American
  Chapter of the Association for Computational Linguistics: Human Language
  Technologies, Volume 1 (Long Papers)}, pages 1112--1122.

\bibitem[{Wu et~al.(2020)Wu, Lin, Karlsson, Lou, and Huang}]{wu2020single}
Qianhui Wu, Zijia Lin, B{\"o}rje Karlsson, Jian-Guang Lou, and Biqing Huang.
  2020.
\newblock Single-/multi-source cross-lingual ner via teacher-student learning
  on unlabeled data in target language.
\newblock In \emph{Proceedings of the 58th Annual Meeting of the Association
  for Computational Linguistics}, pages 6505--6514.

\bibitem[{Wu et~al.(2021)Wu, Ribeiro, Heer, and Weld}]{wu2021polyjuice}
Tongshuang Wu, Marco~Tulio Ribeiro, Jeffrey Heer, and Daniel~S Weld. 2021.
\newblock Polyjuice: Generating counterfactuals for explaining, evaluating, and
  improving models.
\newblock \emph{arXiv preprint arXiv:2101.00288}.

\bibitem[{Wu et~al.(2022)Wu, Gardner, Stenetorp, and Dasigi}]{wu2022generating}
Yuxiang Wu, Matt Gardner, Pontus Stenetorp, and Pradeep Dasigi. 2022.
\newblock Generating data to mitigate spurious correlations in natural language
  inference datasets.
\newblock In \emph{Proceedings of the 60th Annual Meeting of the Association
  for Computational Linguistics (ACL 2022)}.

\bibitem[{Xie et~al.(2020)Xie, Dai, Hovy, Luong, and Le}]{xie2020unsupervised}
Qizhe Xie, Zihang Dai, Eduard Hovy, Thang Luong, and Quoc Le. 2020.
\newblock Unsupervised data augmentation for consistency training.
\newblock \emph{Advances in Neural Information Processing Systems},
  33:6256--6268.

\bibitem[{Xing et~al.(2020)Xing, Jin, Jin, Wang, Zhang, and
  Huang}]{xing2020tasty}
Xiaoyu Xing, Zhijing Jin, Di~Jin, Bingning Wang, Qi~Zhang, and Xuan-Jing Huang.
  2020.
\newblock Tasty burgers, soggy fries: Probing aspect robustness in aspect-based
  sentiment analysis.
\newblock In \emph{Proceedings of the 2020 Conference on Empirical Methods in
  Natural Language Processing (EMNLP)}, pages 3594--3605.

\bibitem[{Yang et~al.(2021)Yang, Li, Cunningham, Zhang, Smyth, and
  Dong}]{yang2021exploring}
Linyi Yang, Jiazheng Li, P{\'a}draig Cunningham, Yue Zhang, Barry Smyth, and
  Ruihai Dong. 2021.
\newblock Exploring the efficacy of automatically generated counterfactuals for
  sentiment analysis.
\newblock In \emph{Proceedings of the 59th Annual Meeting of the Association
  for Computational Linguistics and the 11th International Joint Conference on
  Natural Language Processing (Volume 1: Long Papers)}, pages 306--316.

\bibitem[{Yang et~al.(2022{\natexlab{a}})Yang, Zhang, Qin, Li, Wang, Liu, Wang,
  Xie, and Zhang}]{yang2022glue}
Linyi Yang, Shuibai Zhang, Libo Qin, Yafu Li, Yidong Wang, Hanmeng Liu, Jindong
  Wang, Xing Xie, and Yue Zhang. 2022{\natexlab{a}}.
\newblock Glue-x: Evaluating natural language understanding models from an
  out-of-distribution generalization perspective.
\newblock \emph{arXiv preprint arXiv:2211.08073}.

\bibitem[{Yang et~al.(2022{\natexlab{b}})Yang, Cui, Ning, Wu, and
  Zhang}]{yang-etal-2022-challenges}
Sen Yang, Leyang Cui, Ruoxi Ning, Di~Wu, and Yue Zhang. 2022{\natexlab{b}}.
\newblock \href {https://aclanthology.org/2022.findings-acl.11} {Challenges to
  open-domain constituency parsing}.
\newblock In \emph{Findings of the Association for Computational Linguistics:
  ACL 2022}, pages 112--127, Dublin, Ireland. Association for Computational
  Linguistics.

\bibitem[{Ye et~al.(2021)Ye, Li, Hong, Bai, Chen, Zhou, and Li}]{ye2021ood}
Nanyang Ye, Kaican Li, Lanqing Hong, Haoyue Bai, Yiting Chen, Fengwei Zhou, and
  Zhenguo Li. 2021.
\newblock Ood-bench: Benchmarking and understanding out-of-distribution
  generalization datasets and algorithms.
\newblock \emph{arXiv preprint arXiv:2106.03721}.

\bibitem[{Yu et~al.(2019{\natexlab{a}})Yu, Zhang, Yasunaga, Tan, Lin, Li, Er,
  Li, Pang, Chen et~al.}]{yu2019sparc}
Tao Yu, Rui Zhang, Michihiro Yasunaga, Yi~Chern Tan, Xi~Victoria Lin, Suyi Li,
  Heyang Er, Irene Li, Bo~Pang, Tao Chen, et~al. 2019{\natexlab{a}}.
\newblock Sparc: Cross-domain semantic parsing in context.
\newblock In \emph{Proceedings of the 57th Annual Meeting of the Association
  for Computational Linguistics}, pages 4511--4523.

\bibitem[{Yu et~al.(2019{\natexlab{b}})Yu, Jiang, Dong, and
  Feng}]{yu2019reclor}
Weihao Yu, Zihang Jiang, Yanfei Dong, and Jiashi Feng. 2019{\natexlab{b}}.
\newblock Reclor: A reading comprehension dataset requiring logical reasoning.
\newblock In \emph{International Conference on Learning Representations}.

\bibitem[{Yue and Zhou(2020)}]{yue2020phicon}
Xiang Yue and Shuang Zhou. 2020.
\newblock Phicon: Improving generalization of clinical text de-identification
  models via data augmentation.
\newblock In \emph{Proceedings of the 3rd Clinical Natural Language Processing
  Workshop}, pages 209--214.

\bibitem[{Zellers et~al.(2018)Zellers, Bisk, Schwartz, and
  Choi}]{zellers-etal-2018-swag}
Rowan Zellers, Yonatan Bisk, Roy Schwartz, and Yejin Choi. 2018.
\newblock \href {https://doi.org/10.18653/v1/D18-1009} {{SWAG}: A large-scale
  adversarial dataset for grounded commonsense inference}.
\newblock In \emph{Proceedings of the 2018 Conference on Empirical Methods in
  Natural Language Processing (EMNLP)}, pages 93--104, Brussels, Belgium.
  Association for Computational Linguistics.

\bibitem[{Zhang et~al.(2020)Zhang, Williams, Titov, and
  Sennrich}]{zhang2020improving}
Biao Zhang, Philip Williams, Ivan Titov, and Rico Sennrich. 2020.
\newblock Improving massively multilingual neural machine translation and
  zero-shot translation.
\newblock In \emph{Proceedings of the 58th Annual Meeting of the Association
  for Computational Linguistics}, pages 1628--1639.

\bibitem[{Zhang et~al.(2018)Zhang, Cisse, Dauphin, and
  Lopez-Paz}]{zhang2018mixup}
Hongyi Zhang, Moustapha Cisse, Yann~N Dauphin, and David Lopez-Paz. 2018.
\newblock mixup: Beyond empirical risk minimization.
\newblock In \emph{International Conference on Learning Representations}.

\bibitem[{Zhang et~al.(2019{\natexlab{a}})Zhang, Zhang, Liu, Zhao, Zhu, and
  Chen}]{zhang2019interactive}
Kai Zhang, Hefu Zhang, Qi~Liu, Hongke Zhao, Hengshu Zhu, and Enhong Chen.
  2019{\natexlab{a}}.
\newblock Interactive attention transfer network for cross-domain sentiment
  classification.
\newblock In \emph{Proceedings of the AAAI Conference on Artificial
  Intelligence}, volume~33, pages 5773--5780.

\bibitem[{Zhang et~al.(2019{\natexlab{b}})Zhang, Feng, Meng, You, and
  Liu}]{zhang2019bridging}
Wen Zhang, Yang Feng, Fandong Meng, Di~You, and Qun Liu. 2019{\natexlab{b}}.
\newblock Bridging the gap between training and inference for neural machine
  translation.
\newblock In \emph{Proceedings of the 57th Annual Meeting of the Association
  for Computational Linguistics}, pages 4334--4343.

\bibitem[{Zhang et~al.(2015)Zhang, Zhao, and LeCun}]{zhang2015character}
Xiang Zhang, Junbo Zhao, and Yann LeCun. 2015.
\newblock Character-level convolutional networks for text classification.
\newblock \emph{Advances in neural information processing systems}, 28.

\bibitem[{Zhao et~al.(2019)Zhao, Wang, Yatskar, Cotterell, Ordonez, and
  Chang}]{zhao2019gender}
Jieyu Zhao, Tianlu Wang, Mark Yatskar, Ryan Cotterell, Vicente Ordonez, and
  Kai-Wei Chang. 2019.
\newblock Gender bias in contextualized word embeddings.
\newblock In \emph{Proceedings of NAACL-HLT}, pages 629--634.

\bibitem[{Zhao et~al.(2018{\natexlab{a}})Zhao, Wang, Yatskar, Ordonez, and
  Chang}]{zhao2018gender}
Jieyu Zhao, Tianlu Wang, Mark Yatskar, Vicente Ordonez, and Kai-Wei Chang.
  2018{\natexlab{a}}.
\newblock Gender bias in coreference resolution: Evaluation and debiasing
  methods.
\newblock In \emph{Proceedings of the 2018 Conference of the North American
  Chapter of the Association for Computational Linguistics: Human Language
  Technologies, Volume 2 (Short Papers)}, pages 15--20.

\bibitem[{Zhao et~al.(2018{\natexlab{b}})Zhao, Zhou, Li, Wang, and
  Chang}]{zhao2018learning}
Jieyu Zhao, Yichao Zhou, Zeyu Li, Wei Wang, and Kai-Wei Chang.
  2018{\natexlab{b}}.
\newblock Learning gender-neutral word embeddings.
\newblock In \emph{Proceedings of the 2018 Conference on Empirical Methods in
  Natural Language Processing}, pages 4847--4853.

\bibitem[{Zheng and Lapata(2021)}]{zheng2021disentangled}
Hao Zheng and Mirella Lapata. 2021.
\newblock Disentangled sequence to sequence learning for compositional
  generalization.
\newblock \emph{arXiv preprint arXiv:2110.04655}.

\bibitem[{Zheng et~al.(2020)Zheng, Chen, and Huang}]{zheng2020out}
Yinhe Zheng, Guanyi Chen, and Minlie Huang. 2020.
\newblock Out-of-domain detection for natural language understanding in dialog
  systems.
\newblock \emph{IEEE/ACM Transactions on Audio, Speech, and Language
  Processing}, 28:1198--1209.

\bibitem[{Zhong et~al.(2020)Zhong, Xiao, Tu, Zhang, Liu, and
  Sun}]{zhong2020does}
Haoxi Zhong, Chaojun Xiao, Cunchao Tu, Tianyang Zhang, Zhiyuan Liu, and Maosong
  Sun. 2020.
\newblock How does nlp benefit legal system: A summary of legal artificial
  intelligence.
\newblock In \emph{Proceedings of the 58th Annual Meeting of the Association
  for Computational Linguistics}, pages 5218--5230.

\bibitem[{Zhong et~al.(2021{\natexlab{a}})Zhong, Lee, Zhang, and
  Klein}]{Zhong2021AdaptingLM}
Ruiqi Zhong, Kristy Lee, Zheng Zhang, and Dan Klein. 2021{\natexlab{a}}.
\newblock Adapting language models for zero-shot learning by meta-tuning on
  dataset and prompt collections.
\newblock In \emph{EMNLP}.

\bibitem[{Zhong et~al.(2021{\natexlab{b}})Zhong, Wang, Tang, Xu, Guo, Wang,
  Yin, Zhou, and Duan}]{zhong2021ar}
Wanjun Zhong, Siyuan Wang, Duyu Tang, Zenan Xu, Daya Guo, Jiahai Wang, Jian
  Yin, Ming Zhou, and Nan Duan. 2021{\natexlab{b}}.
\newblock Ar-lsat: Investigating analytical reasoning of text.
\newblock \emph{arXiv preprint arXiv:2104.06598}.

\bibitem[{Zhou et~al.(2021{\natexlab{a}})Zhou, Levy, Li, Ghazvininejad, and
  Neubig}]{zhou2021distributionally}
Chunting Zhou, Daniel Levy, Xian Li, Marjan Ghazvininejad, and Graham Neubig.
  2021{\natexlab{a}}.
\newblock Distributionally robust multilingual machine translation.
\newblock In \emph{Proceedings of the 2021 Conference on Empirical Methods in
  Natural Language Processing}, pages 5664--5674.

\bibitem[{Zhou et~al.(2021{\natexlab{b}})Zhou, He, Li, Bing, Cambria, Si, and
  Miao}]{zhou2021melm}
Ran Zhou, Ruidan He, Xin Li, Lidong Bing, Erik Cambria, Luo Si, and Chunyan
  Miao. 2021{\natexlab{b}}.
\newblock Melm: Data augmentation with masked entity language modeling for
  cross-lingual ner.
\newblock \emph{arXiv preprint arXiv:2108.13655}.

\bibitem[{Zhou et~al.(2021{\natexlab{c}})Zhou, Liu, and
  Chen}]{zhou2021contrastive}
Wenxuan Zhou, Fangyu Liu, and Muhao Chen. 2021{\natexlab{c}}.
\newblock Contrastive out-of-distribution detection for pretrained
  transformers.
\newblock \emph{arXiv preprint arXiv:2104.08812}.

\bibitem[{Zhu et~al.(2019)Zhu, Cheng, Gan, Sun, Goldstein, and
  Liu}]{zhu2019freelb}
Chen Zhu, Yu~Cheng, Zhe Gan, Siqi Sun, Tom Goldstein, and Jingjing Liu. 2019.
\newblock Freelb: Enhanced adversarial training for natural language
  understanding.
\newblock In \emph{International Conference on Learning Representations}.

\bibitem[{Zhu and Goldberg(2009)}]{zhu2009introduction}
Xiaojin Zhu and Andrew~B Goldberg. 2009.
\newblock Introduction to semi-supervised learning.
\newblock \emph{Synthesis lectures on artificial intelligence and machine
  learning}, 3(1):1--130.

\bibitem[{Zhu et~al.(2021)Zhu, Balagopalan, Ghassemi, and
  Rudzicz}]{zhu2021quantifying}
Zining Zhu, Aparna Balagopalan, Marzyeh Ghassemi, and Frank Rudzicz. 2021.
\newblock Quantifying the task-specific information in text-based
  classifications.
\newblock \emph{arXiv preprint arXiv:2110.08931}.

\bibitem[{Zmigrod et~al.(2019)Zmigrod, Mielke, Wallach, and
  Cotterell}]{zmigrod2019counterfactual}
Ran Zmigrod, Sebastian~J Mielke, Hanna Wallach, and Ryan Cotterell. 2019.
\newblock Counterfactual data augmentation for mitigating gender stereotypes in
  languages with rich morphology.
\newblock In \emph{Proceedings of the 57th Annual Meeting of the Association
  for Computational Linguistics}, pages 1651--1661.

\end{thebibliography}
\bibliographystyle{acl_natbib}

\clearpage
\onecolumn

\appendix
\section{Appendix}\label{app:full_table}
% \begin{figure}[h]
% 	\centering
% % 	\setlength{\belowcaptionskip}{-0.5cm} 
% 	\includegraphics[width=\linewidth]{mindmap2.pdf}
% 	\caption{Overview of OOD-related Research in NLP.}
% 	\label{fig:scope}
% \end{figure}

\begin{table*}[h]
    \centering
    \small
    \resizebox{\linewidth}{!}{
\begin{tabular}{@{}l|ccccc@{}}
\toprule
\multicolumn{1}{c|}{Work} & Task & Scope         & Method    & Dataset  & Metric  \\ \midrule
\citet{dong2016language}   &\begin{tabular}[c]{@{}c@{}}MRC \end{tabular}      & \begin{tabular}[c]{@{}c@{}}Logical \\ Reasoning \\(Domain Variance) \end{tabular} &  \parbox[c]{6cm}{Propose an attention-enhanced encoder-decoder model  invariant representation}  & \parbox[c]{6cm}{JOBS, GEO, ATIS, IFTTT}  & Accuracy        \\ \midrule
\citet{yu2019reclor}   &\begin{tabular}[c]{@{}c@{}}MRC \end{tabular}      & \begin{tabular}[c]{@{}c@{}}Logical \\ Reasoning (Bias) \end{tabular} &  \parbox[c]{6cm}{Introduce a new Reading Comprehension dataset requiring logical reasoning (ReClor) extracted from standardized graduate admission examinations.}  & \parbox[c]{6cm}{ReClor}  & Accuracy        \\ \midrule
\citet{liu2021logiqa}   &\begin{tabular}[c]{@{}c@{}}MRC \end{tabular}      & \begin{tabular}[c]{@{}c@{}}Logical \\ Reasoning (Bias) \end{tabular} &  \parbox[c]{6cm}{Introduce a comprehensive dataset which is sourced from expert-written questions.}  & \parbox[c]{6cm}{Logiqa}  & Accuracy        \\ \midrule
\citet{zhong2021ar}   & \begin{tabular}[c]{@{}c@{}}MRC \end{tabular}     & \begin{tabular}[c]{@{}c@{}}Logical \\ Reasoning (Bias) \end{tabular} &  \parbox[c]{6cm}{Introduce a new dataset consisting of questions from the Law School Admission Test from 1991 to 2016.}  & \parbox[c]{6cm}{AR-LSAT}  & Accuracy        \\ \midrule
\citet{kaushik2018much}   & MRC     & \begin{tabular}[c]{@{}c@{}} Annotation artifacts \end{tabular}&  \parbox[c]{6cm}{Establish sensible baselines for the bAbI, SQuAD, CBT, CNN, and Who-did-What datasets, finding that question- and passage-only models often perform surprisingly well.}  & \parbox[c]{6cm}{bAbI, SQuAD, CBT, CNN, Whodid-What}  & Accuracy        \\ \midrule
\citet{sugawara2018makes}   & MRC     & \begin{tabular}[c]{@{}c@{}} Annotation artifacts\end{tabular}&  \parbox[c]{6cm}{Establish sensible baselines for the bAbI, SQuAD, CBT, CNN, and Whodid-What datasets, finding that question- and passage-only models often perform surprisingly well.}  & \parbox[c]{6cm}{QA4MRE, CNN/Daily Mail, Children's Book, WikiReading, LAMBADA, Who-did-What, ProPara, CliCR, SQuAD, DuoRC}  & Accuracy        \\ \midrule
\citet{sugawara2020assessing} & MRC     & \begin{tabular}[c]{@{}c@{}}Annotation artifacts\\(shortcut) \end{tabular}&  \parbox[c]{6cm}{Propose a semi-automated, ablation-based methodology to evaluate capacity of MRC datasets.}  & \parbox[c]{6cm}{CoQA,DuoRC, HotpotQA, SquAD, SQuAD, ARC, MCTest, MultiRC, RACE, SWAG}  & Accuracy  F1      \\ \midrule
\citet{bartolo2020beat}  & MRC     & \begin{tabular}[c]{@{}c@{}}Annotation artifacts\\(shortcut) \end{tabular}&  \parbox[c]{6cm}{Propose an adversarial annotation data collection method.  Training on adversarially collected samplesleads to strong generalization.}  & \parbox[c]{6cm}{SQuAD1.1}  & F1      \\ \midrule
\citet{lai2021machine}  & MRC     & \begin{tabular}[c]{@{}c@{}}Annotation artifacts\\(shortcut) \end{tabular}&  \parbox[c]{6cm}{Propose two synthetic dataset and two new method to investigate shortcut in MRC especially on paraphrasing.}  & \parbox[c]{6cm}{QWM-Para  dataset derived from SQuAD}  & F1      \\ \midrule
\citet{cheng-etal-2019-robust}   & \begin{tabular}[c]{@{}c@{}} NLG \end{tabular}     & \begin{tabular}[c]{@{}c@{}} Data noise \end{tabular}&  \parbox[c]{6cm}{Propose double adversarial input MT model to improve the robustness.}  & \parbox[c]{6cm}{LDC corpus, NIST, WMT’14, newstest2013,2014}  & BLEU score      \\ \midrule
\citet{zhang2019bridging}   & \begin{tabular}[c]{@{}c@{}} NLG \end{tabular}     & \begin{tabular}[c]{@{}c@{}} Annotation artifacts\\(exposure bias) \end{tabular}&  \parbox[c]{6cm}{In word-level sampling context words is not only from the ground truth sequence but also from the predicted sequence by the model during training, where the predicted sequence is selected with a sentence-level optimum.}  & \parbox[c]{6cm}{ NIST, WMT’14}  & BLEU score      \\ \midrule
\citet{zhou2021distributionally}   & \begin{tabular}[c]{@{}c@{}} NLG \end{tabular}     & \begin{tabular}[c]{@{}c@{}} Annotation artifacts\\(domain) \end{tabular}&  \parbox[c]{6cm}{Propose a new learning objective for MNMT based on DRO.}  & \parbox[c]{6cm}{58-languages TED talk corpus, WMT}  & \begin{tabular}[c]{@{}c@{}} BLEU score\end{tabular}     \\ \midrule
\citet{hewitt2021ensembles}   & \begin{tabular}[c]{@{}c@{}} NLG \end{tabular}     & \begin{tabular}[c]{@{}c@{}} Annotation artifacts\\(domain) \end{tabular}&  \parbox[c]{6cm}{Present methods to combine the benefits of full and lightweight finetuning, achieving strong performance both ID and OOD.}  & \parbox[c]{6cm}{WebNLG, XSUM, Open-domain QA}  & \begin{tabular}[c]{@{}c@{}} BLEU score \\ ROUGE-2 score \\Exact match accuracy\end{tabular} \\  \bottomrule
\end{tabular}
}
    \caption{Methods towards OOD generalization challenge in the task of MRC and NLG.}
    \label{table:full_1}
\end{table*}

\begin{table*}[t]
    \centering
    \small
    \resizebox{\linewidth}{!}{
\begin{tabular}{@{}l|ccccc@{}}
\toprule
\multicolumn{1}{c|}{Work} & Task & Scope         & Method    & Dataset  & Metric  \\ \midrule
\citet{jia2019cross}     & NER     & Input variance & \parbox[c]{6cm}{Design cross-domain and cross-
task network for NER domain generalization.}  & \parbox[c]{6cm}{CoNLL, BioNLP13PC, BioNKP13CG,\\ CBS Newws}  & F1           \\ \midrule
\citet{jia2020multi}     & NER     & Input variance & \parbox[c]{6cm}{Multi-task learning with multi-cell LSTM for NER domain generalization.}  & \parbox[c]{6cm}{CoNLL2003, Broad Twitter, Twitter, BioNLP13PC, BioNLP13CG, CBS News} & F1         \\ \midrule
\citet{liu2021crossner}  & NER     & Input variance & \parbox[c]{6cm}{Introduce  a cross-domain NER dataset with a domain-related corpus and propose a baseline.}  & \parbox[c]{6cm}{CoNLL2003, CrossNER}     & F1     \\ \midrule
\citet{chen2021data}     & NER     & Input variance &  \parbox[c]{6cm}{Data Augmentation for crossdomain NER. Propose a novel neural architecture to transform the data representation from a high-resource to a low-resource domain.}  & \parbox[c]{6cm}{Ontonotes 5.0, Temporal Twitter }  & F1        \\ \midrule
\citet{ghaddar2017winer}     & NER     & Output variance &  \parbox[c]{6cm}{Propose a large, high quality, annotated corpus WiNER for cross-domain NER.}  & \parbox[c]{6cm}{CoNLL, MUC, ONTO, WGOLD, WEB}  & F1        \\ \midrule 
\citet{vu2020effective}     & NER     & Output variance &  \parbox[c]{6cm}{Adversarially trained masked LMs with domain generalization.}  & \parbox[c]{6cm}{CoNLL2003, WNUT2016, FIN, JNLPBA, BC2GM, BioNLP09, BioNLP11EPI}  & F1        \\ \midrule
\citet{wu2020single}     & NER     & Output variance &  \parbox[c]{6cm}{Propose a teacher-student learning method fro cross-linguial NER.}  & \parbox[c]{6cm}{CoNLL-2002, CoNLL-2003}  & F1        \\ \midrule
\citet{Nguyen2021DOZENCZ}     & NER     & Output variance &  \parbox[c]{6cm}{Cross domain zero shot NER with knowledge base.}  & \parbox[c]{6cm}{music, science datset}  & F1        \\ \midrule
\citet{cui2021template}     & NER     & Output variance &  \parbox[c]{6cm}{Propose a template-based method for NER, treating NER as a language model ranking problem in a sequence-to-sequence framework, where original sentences and statement templates filled by candidate named entity span are regarded as the source sequence and the target sequence.}  & \parbox[c]{6cm}{CoNLL, MIT Movie Review, MIT Restaurant Review}  & F1        \\ \midrule
\citet{ma2021template}     & NER     & Output variance &  \parbox[c]{6cm}{Reformulate NER tasks as LM problems without templates.}  & \parbox[c]{6cm}{CoNLL2003, Ontonotes 5.0, MIT-Movie}  & F1        \\ \midrule
\citet{zhou2021melm}     & NER     & Output variance &  \parbox[c]{6cm}{Propose Masked Entity Language Modeling (MELM) as a novel data augmentation framework for low-resource NER to alleviate the token-label misalignment.}  & \parbox[c]{6cm}{CoNLL}  & F1        \\ \midrule
\citet{lee2021good}     & NER     & Output variance &  \parbox[c]{6cm}{Propose a  demonstration-based learning method for NER, which lets the input be prefaced by task demonstrations for in-context learning.}  & \parbox[c]{6cm}{CoNLL-2003, Ontonotes 5.0, BC5CDR}  & F1        \\ \midrule
\citet{das2021container}     & NER     & Output variance &  \parbox[c]{6cm}{Propose  a novel contrastive learning technique that optimizes the inter-token distribution distance instead of class-specific attributes for Few-Shot NER.}  & \parbox[c]{6cm}{OntoNotes, CoNLL’03, WNUT ’17, GUM}  & F1        \\ \midrule
\citet{wang2022miner}   & NER     & Output variance &  \parbox[c]{6cm}{Propose an information theoretic perspective method to imporve  out-of-vocabulary entities prediction.}  & \parbox[c]{6cm}{WNUT2017,TwitterNER,BioNER, Conll03-Typos, Conll03-OOV}  & F1        \\ \midrule
\citet{liu2021noisy}   & NER     & Data noise &  \parbox[c]{6cm}{Propose a calibrated confidence estimation and integrate it into a self-training framework for boosting performance in general noisy settiings.}  & \parbox[c]{6cm}{CoNLL, Tweet, Webpage, Wikigold}  & F1        \\ \midrule
\citet{gu2021beyond}   & QA     & \begin{tabular}[c]{@{}c@{}} Compositional \\generlization (Bias) \end{tabular}&  \parbox[c]{6cm}{Construct  new large-scale, high-quality dataset GrailQA, and propose a novel BERT-based KBQA model.}  & \parbox[c]{6cm}{GRAILQA}  & F1        \\ \midrule
\citet{lewis2021question}   & QA     & \begin{tabular}[c]{@{}c@{}} Compositional \\generlization (Bias) \end{tabular}&  \parbox[c]{6cm}{Evaluate three popular open-domain benchmark datasets and  find that all models perform dramatically worse on questions that cannot be memorized from training sets.}  & \parbox[c]{6cm}{WebQuestions, TriviaQA, Open Natural Questions}  & Exact match score        \\ \midrule
\citet{bogin2021latent}   & QA     & \begin{tabular}[c]{@{}c@{}} Compositional \\generlization (Bias) \end{tabular}&  \parbox[c]{6cm}{Propose a model that computes a representation and denotation for all question spans in a bottom-up, compositional manner using a CKY-style parser. Inductive bias towards tree structures dramatically improves systematic generalization to out-of-distribution examples.}  & \parbox[c]{6cm}{arithmetic expressions benchmark, CLEVR, CLOSURE}  & F1        \\ \midrule
\citet{cai2017pay}   & QA     & \begin{tabular}[c]{@{}c@{}} Compositional \\ generalization \end{tabular}&  \parbox[c]{6cm}{Propose a hierarchical RNN with attention to encode the sentence in the story and score candidate endings.}  & \parbox[c]{6cm}{ROC Story}  & Accuracy        \\ \midrule
\citet{min2019compositional}   & QA     & \begin{tabular}[c]{@{}c@{}} Compositional \\ generalization \end{tabular}&  \parbox[c]{6cm}{Propose  a single-hop BERT-based RC model.}  & \parbox[c]{6cm}{HOTPOTQA}  & F1        \\ \midrule
\citet{bartolo2021models}   & QA     & \begin{tabular}[c]{@{}c@{}} Compositional \\ generalization \\ (domain) \end{tabular}&  \parbox[c]{6cm}{Introduce a generator-in-the-loop model to provide real-time suggestions for annotator, which maintains the advantages of DADC and reduce annotation cost.}  & \parbox[c]{6cm}{SQuAD1.1, AdversarialQA, GAA-assisted data}  & \parbox[c]{3cm}{ Median time per example \\ Validated Model Error Rate (vMER) \\Median time per validated model-fooling example \\ Downstream effectiveness (F1 score)}      \\ \midrule
\citet{lyu2022extending}   & QA     & \begin{tabular}[c]{@{}c@{}} Compositional \\ generalization\\(domain) \end{tabular}&  \parbox[c]{6cm}{Extend the scope of “OOD” by splitting QA examples into different subdomains according to their several internal characteristics including question type, text length, answer position. Examine the performance of QA systems trained on the data from different subdomains.}  & \parbox[c]{6cm}{SQuAD 1.1, NewsQA}  & F1        \\   \bottomrule
\end{tabular}
}
    \caption{Methods towards OOD generalization challenge in the task of NER and QA.}
    \label{table:full_2}
\end{table*}

\begin{table*}[]
    \centering
    \small
    \resizebox{\linewidth}{!}{
\begin{tabular}{@{}l|ccccc@{}}
\toprule
\multicolumn{1}{c|}{Work} & Task & Scope         & Method    & Dataset   & Metric  \\ \midrule
\citet{wang2021identifying}   & \begin{tabular}[c]{@{}c@{}} SA \end{tabular}     & \begin{tabular}[c]{@{}c@{}} Annotation artifacts \\(shortcut) \end{tabular}&  \parbox[c]{6cm}{Automatically identify such spurious correlations in NLP models at scale.}  & \parbox[c]{6cm}{SST, Yelp, Occupation dataset, Amazon Kitchen, Amazon Electronics}  &\begin{tabular}[c]{@{}c@{}}  Precision\\ Importance score    \end{tabular}    \\ \midrule
\citet{kaushik2019learning}         & SA, NLI     & Input variance  & CDA.  & \parbox[c]{6cm}{SNLI, IMDB}  & Accuracy           \\ \midrule
\citet{kaushik2020explaining}     & SA, NLI     & Input variance  & evaluate the efficacy of CDA.  & \parbox[c]{6cm}{IMBb, Yelp, Amazon, Semeval, CRD, SNLI, MultiNLI} & Accuracy         \\ \midrule
\citet{hendrycks2020pretrained}     & SA, NLI     & Input variance  & \parbox[c]{6cm}{\centering evaluate OOD generalization of pre-trained model.}  & \parbox[c]{6cm}{SST-2,Yelp Review,Amazon Review,MultiNLI} & Accuracy         \\ \midrule
\citet{Wang2020IdentifyingSC}       & SA            & Input variance     & \parbox[c]{6cm}{Train spurious feature detector \& improve OOD generalization.}  & \parbox[c]{6cm}{IMDB reviews, Kindle reviews, Toxic comment,Toxic tweet}  & Accuracy        \\ \midrule
\citet{wang2021robustness}       & SA            & Input variance     & \parbox[c]{6cm}{train spurious feature detector \& Improve robustness to spurious correlations via CDA.}  & \parbox[c]{6cm}{IMDB reviews, Amazon, Kindle reviews}  & Accuracy        \\ \midrule
\citet{yang2021exploring}       & SA            & Input variance     & \parbox[c]{6cm}{CDA \& improve OOD genralization.}  & \parbox[c]{6cm}{SST-2, IMDB, Amazon Reviews, Semeval 2017, Yelp Reviews}  & Accuracy        \\ \midrule
\citet{lu2022rationale}       & SA            & Input variance     & \parbox[c]{6cm}{improving robustness via auto Semi-factual data augmentation}  & \parbox[c]{6cm}{IMDb, Amazon reviews, Yelp reviews, SST, SemEval-2017 Twitter.}  & Accuracy        \\ \midrule
\citet{chen2018multinomial}       & SA            & Data noise     & \parbox[c]{6cm}{improving OOD generalization via learning invariant features.}  & \parbox[c]{6cm}{Amazon reviews, FDU-MTL dataset}  & Accuracy        \\ \midrule
\citet{Johnson2017GooglesMN}       & MT            & Output variance     & \parbox[c]{6cm}{zero-shot MT via training on multilingual corpus}  & \parbox[c]{6cm}{WMT’14, WMT’15.}  & BLEU score      \\ \midrule
\citet{zhang2020improving}       & MT            & Output variance     & \parbox[c]{6cm}{improve zero-shot MT: enforce translation to the target language via backtranslation.}  & \parbox[c]{6cm}{OPUS-100}  & \begin{tabular}[c]{@{}c@{}} BLEU score\\ Win ratio \end{tabular}       \\ \midrule
\citet{arivazhagan2019missing}       & MT            & Output variance     & \parbox[c]{6cm}{improve zero-shot MT: learn invariant representations via auxiliary losses.}  & \parbox[c]{6cm}{newstest-2012, WMT14, newstest-2013, IWSLT 2017}  & BLEU score       \\ \midrule
\citet{ji2020cross}       & MT            & Output variance     & \parbox[c]{6cm}{improve zero-shot MT: obtain an universal encoder for different languages.}  & \parbox[c]{6cm}{Europarl, MultiUN}  & BLEU score       \\ \midrule
\citet{liu2021improving}       & MT            & Output variance     & \parbox[c]{6cm}{improve zero-shot MT: removing residual connections.}  & \parbox[c]{6cm}{IWSLT 2017, Europarl v7, PMIndia}  & BLEU score       \\ \midrule
\citet{zheng2021disentangled}       & MT            & \begin{tabular}[c]{@{}c@{}}  Compositional\\ generalization \end{tabular} & \parbox[c]{6cm}{Improve composional generalization: propose an extension to sequence-to-sequence models which encourages disentanglement.}  & \parbox[c]{6cm}{COGS, CFQ}  & \begin{tabular}[c]{@{}c@{}} BLEU score\\ Exact match score\\ Compound translation error rate \end{tabular}
        \\ \midrule
\citet{belinkov2017synthetic}       & MT            & Data noise
     & \parbox[c]{6cm}{Increase model robustness: structure-invariant word representations \& robust training.}  & \parbox[c]{6cm}{IWSLT 2016, WiCoPaCo, Wikipedia Revision Dataset, The MERLIN corpus, Czech: manually annotated essays}  & BLEU score
        \\ \midrule
\citet{stanovsky2019evaluating}       & MT            & Annotation artifacts
     & \parbox[c]{6cm}{present the challenge set for evaluating gender bias in machine tranlation.}  & \parbox[c]{6cm}{WinoMT}  & \begin{tabular}[c]{@{}c@{}} Accuracy\\ F1 \end{tabular}
        \\ \midrule
\citet{choubey2021improving}       & MT            & Annotation artifacts
     & \parbox[c]{6cm}{propose gender-filtered self-training (GFST) to improve gender translation accuracy.}  & \parbox[c]{6cm}{WinoMT, MuST-SHE}  & \begin{tabular}[c]{@{}c@{}} Accuracy\\ F1\\ Recall\\ BLEU score \end{tabular}
        \\ \midrule
\citet{williams2018broad}       & NLI            & Input variance & \parbox[c]{6cm}{introduce MultiNLI benchmark.}  & \parbox[c]{6cm}{MultiNLI, SNLI}  & Accuracy
        \\ \midrule
\citet{naik2018stress}       & NLI            & Annotation artifacts &  \parbox[c]{6cm}{propose Stress Test dataset for NLI.}  & \parbox[c]{6cm}{MultiNLI}  & \begin{tabular}[c]{@{}c@{}} Accuracy\\ Error rate \end{tabular}
        \\ \midrule        
\citet{zellers-etal-2018-swag}       & NLI            & Annotation artifacts &  \parbox[c]{6cm}{propose dataset SWAG for measuring common reasoning of NLI model.}  & \parbox[c]{6cm}{SWAG, SNLI}  & Accuracy
        \\ \midrule  
\citet{feng2019misleading}       & NLI            & Annotation artifacts &  \parbox[c]{6cm}{We illustrate how partial-input baselines can overshadow trivial.}  & \parbox[c]{6cm}{SNLI}  & Accuracy
        \\ \midrule          
\citet{mccoy2019right}       & NLI            & Annotation artifacts &  \parbox[c]{6cm}{Introduced HANS dataset which contains three fallible syntactic heuristics.}  & \parbox[c]{6cm}{MultiNLI, HANS}  & Accuracy
        \\ \midrule                  
\citet{le2020adversarial}       & NLI            & Annotation artifacts &  \parbox[c]{6cm}{Use AFLITE to reduce dataset biases, thus improve OOD generalization.}  & \parbox[c]{6cm}{SNLI, ANLI, HANS, NLI-Diagnostics, Stress tests, QNLI, MultiNLI}  & Accuracy
        \\ \midrule           
\citet{sakaguchi2020winogrande}       & NLI            & Annotation artifacts &  \parbox[c]{6cm}{Introduce WINOGRANDE, which is harder \& larger than Winograd Schema Challenge.}  & \parbox[c]{6cm}{WINOGRANDE, WSC, DPR , COPA, KnowRef, Winogender}  & Accuracy
        \\ \midrule                   
\citet{nie-etal-2020-adversarial}       & NLI            & Annotation artifacts &  \parbox[c]{6cm}{Introduce ANLI, collected via iterative\&adversarial human-and-model-in-the-loop procedure.}  & \parbox[c]{6cm}{ANLI, SNLI, MultiNLI, SNLI-Hard, NLI Stress Tests}  &  \begin{tabular}[c]{@{}c@{}} Accuracy\\ Error rate \end{tabular}
        \\ \midrule         
\citet{liu-etal-2020-hyponli}       & NLI            & Annotation artifacts &  \parbox[c]{6cm}{derive adversarial examples in terms of the hypothesis-only bias and explore eligible ways to mitigate such bias.}  & \parbox[c]{6cm}{SNLI, MultiNLI}  & Accuracy
        \\ \midrule               
\citet{wu2022generating}       & NLI            & Annotation artifacts &  \parbox[c]{6cm}{generating debiased datasets through filter out instances contribute to spurious correlations.}  & \parbox[c]{6cm}{SNLI, MultiNLI, HANS, SNLI-hard, MultiNLI-hard}  & Accuracy
        \\  \midrule                 
\citet{du2021towards}   & \begin{tabular}[c]{@{}c@{}} NLI \end{tabular}     & \begin{tabular}[c]{@{}c@{}} Annotation artifacts \\(shortcut) \end{tabular}&  \parbox[c]{6cm}{Propose a shortcut mitigation framework LTGR using knowledge distillation framework, to suppress the model from making overconfident predictions for samples with large shortcut degree.}  & \parbox[c]{6cm}{MultiNLI, FEVER, and MultiNLI-backdoor }  & Accuracy      \\ \midrule
\citet{liu2021just}   & \begin{tabular}[c]{@{}c@{}} NLI \end{tabular}     & \begin{tabular}[c]{@{}c@{}} Annotation artifacts \\(domain)\end{tabular}&  \parbox[c]{6cm}{Propose a simple two-stage approach, that minimizes the loss over a reweighted dataset (second stage) where we upweight training examples that are misclassified at the end of a few steps of standard training (first stage). It overcome the requirement of expensive group annotations in group DRO.}  & \parbox[c]{6cm}{MultiNLI, CivilComments-WILDS}  & Accuracy      \\     \midrule
\citet{oren2019distributionally}   & \begin{tabular}[c]{@{}c@{}} Text \\classification \end{tabular}     & \begin{tabular}[c]{@{}c@{}} Annotation artifacts\\ (bias) \end{tabular}&  \parbox[c]{6cm}{Propose a new DRO based approach called topic conditional value at risk.}  & \parbox[c]{6cm}{Yelp, ONEBWORD,  TPIPADV}  & perplexity        \\ \bottomrule
\end{tabular}
}
    \caption{Methods towards OOD generalization challenge in the task of SA, NLI, and MT.}
    \label{table:full_3}
\end{table*}

\end{document}